\documentclass[9pt,twocolumn,twoside]{pnas-new}

\templatetype{pnasbriefreport}

\fancypagestyle{firststyle}{
  \fancyfoot[R]{\footerfont \textbf{\today}\hspace{7pt}|\hspace{7pt}\textbf{\thepage}}
  \fancyfoot[L]{}
}
\usepackage[capitalise]{cleveref}
\usepackage{overpic}
\usepackage{amsthm}
\usepackage{mathrsfs}  
\newtheorem{theorem}{Theorem}

\newcommand{\R}{\mathbb{R}}
\newcommand{\F}{\mathscr{F}}

\newcommand{\HS}{\textup{HS}}
\renewcommand{\d}{\,\textup{d}}
\renewcommand{\L}{\mathcal{L}}

\usepackage{xcolor}

\usepackage{pdfpages}

\title{Elliptic PDE learning is provably data-efficient}

\author[a,1]{Nicolas Boull\'e}
\author[b]{Diana Halikias} 
\author[b]{Alex Townsend}

\affil[a]{Isaac Newton Institute for Mathematical Sciences, University of Cambridge, Cambridge CB3 0EH, United Kingdom}
\affil[b]{Mathematics Department, Cornell University, Ithaca, NY 14853-4201, United States}

\leadauthor{Boull\'e} 

\authorcontributions{Author contributions: N.B., D.H., and A.T. designed research; performed research; analyzed data; and wrote the paper.}
\authordeclaration{The authors declare no competing interest.}
\correspondingauthor{\textsuperscript{1}To whom correspondence should be addressed. Email: nb690@cam.ac.uk.}

\keywords{deep learning $|$ inverse problems $|$ sample complexity $|$ neural operators} 

\begin{abstract}
PDE learning is an emerging field that combines physics and machine learning to recover unknown physical systems from experimental data. While deep learning models traditionally require copious amounts of training data, recent PDE learning techniques achieve spectacular results with limited data availability. Still, these results are empirical. Our work provides theoretical guarantees on the number of input-output training pairs required in PDE learning. Specifically, we exploit randomized numerical linear algebra and PDE theory to derive a provably data-efficient algorithm that recovers solution operators of 3D uniformly elliptic PDEs from input-output data and achieves an exponential convergence rate of the error with respect to the size of the training dataset with an exceptionally high probability of success.
\end{abstract}

\dates{This manuscript was compiled on \today.}
\doi{}

\begin{document}

\maketitle
\thispagestyle{firststyle}
\ifthenelse{\boolean{shortarticle}}{\ifthenelse{\boolean{singlecolumn}}{\abscontentformatted}{\abscontent}}{}

\dropcap{M}any scientific breakthroughs have come from deriving new partial differential equations (PDEs) from first principles to model real-world phenomena and simulating them on a computer to make predictions. However, many crucial problems currently lack an adequate mathematical formulation. It is not clear how to derive PDEs to describe how turbulence sheds off the wing of a hypersonic aircraft, how E.~coli bacteria swim in unison to form an active fluid, or how atomic particles behave with long-range interactions. Rather than working from first principles, scientists are now looking to derive PDEs from real-world data using deep learning techniques~\cite{karniadakis2021physics}.

The success of deep learning in language models, visual object recognition, and drug discovery is well known~\cite{lecun2015deep}. The emerging field of PDE learning hopes to extend this to discovering new physical laws by supplying deep learning models with experimental or observational data~\cite{karniadakis2021physics,raissi2020hidden}. PDE learning commonly seeks to recover features such as symmetries, conservation laws, solution operators, and the parameters of a family of hypothesized PDEs. In most deep-learning applications, a large amount of data is needed, which is often unrealistic in engineering and biology. However, PDE learning can be shockingly data-efficient in practice~\cite{lu2021learning}. In particular, surprisingly little data is used to learn the solution operator, which maps the forcing term to the solution of the PDE.

In this paper, we provide a theoretical explanation of this behavior by showing that, for $\epsilon>0$ sufficiently small, one can recover an $\epsilon$-approximation to the solution operator of a three-dimensional (3D) elliptic PDE with a training dataset of size about $\mathcal{O}(\log^5(1/\epsilon))$. Elliptic PDEs, such as the steady state heat equation, are ubiquitous in physics and model diffusion phenomena. Solution operators can produce surrogate data for data-intensive machine learning approaches such as learning reduced order models for design optimization in engineering, uncovering physics in climate models, and PDE recovery~\cite{karniadakis2021physics}.

\begin{figure}[t!]
    \centering
    \begin{overpic}[width=8.6cm]{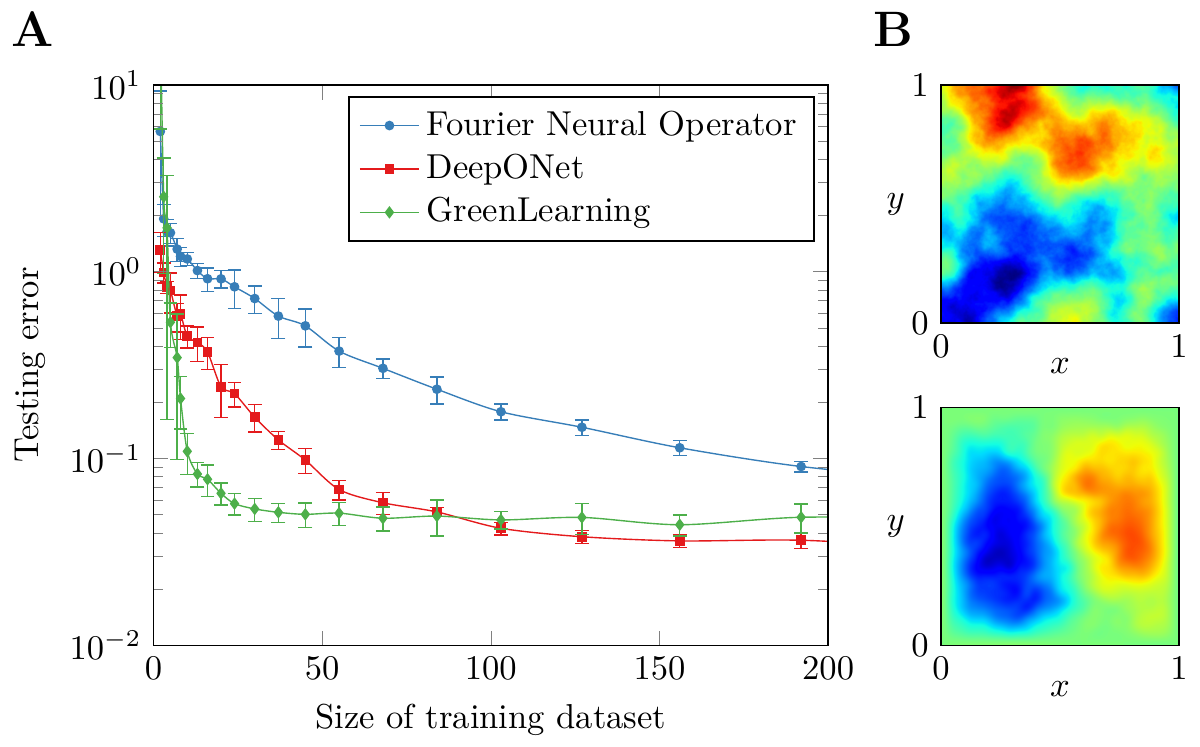}
    \end{overpic}
    \caption{Elliptic PDE learning methods can be data-efficient. (A) Performance of three deep learning techniques in approximating the solution operator of the 2D Poisson equation with zero Dirichlet boundary condition on the domain $[0,1]^2$. On small datasets, DeepONet and GreenLearning attain exponential decay of the testing error, while Fourier Neural Operator (FNO) attains algebraic decay. (B) A forcing term (top) and corresponding predicted solution (bottom) to the 2D Poisson equation by a FNO.}
    \label{fig_exp}
\end{figure}

To illustrate the observed data-efficiency of PDE learning, we compare the performance of three techniques~\cite{li2020fourier,lu2021learning,boulle2022data} for recovering the solution operator associated with the 2D Poisson equation in~\cref{fig_exp}. We vary the size of the training dataset, consisting of random forcing terms and corresponding solutions obtained by a numerical solver. We then evaluate the accuracy of the predicted solutions on a testing dataset with new forcing terms. The three methods are based on deep learning and differ in their neural network architectures. While the Fourier Neural Operator~\cite{li2020fourier} exploits the fast Fourier transform for computationally efficient training, DeepONet~\cite{lu2021learning} and GreenLearning~\cite{boulle2022data} achieve a faster convergence rate on small training datasets. Here, DeepONet employs a complex network architecture with many parameters. In contrast, GreenLearning leverages prior knowledge that the solution operator is an integral operator and the approximation power of rational neural networks~\cite{boulle2020rational}.
Green's function learning is observed to be the most data-efficient in~\cref{fig_exp}, as for a fixed training dataset size, it achieves the smallest testing error. All methods plateau due to discretization errors, and the training procedure gets stuck in a local minimum of the loss landscape rather than finding the global minimum. The rapid decay of testing errors prior to the plateau motivates our main result.

\begin{figure*}[t!]
    \centering
    \begin{overpic}[width=\textwidth]{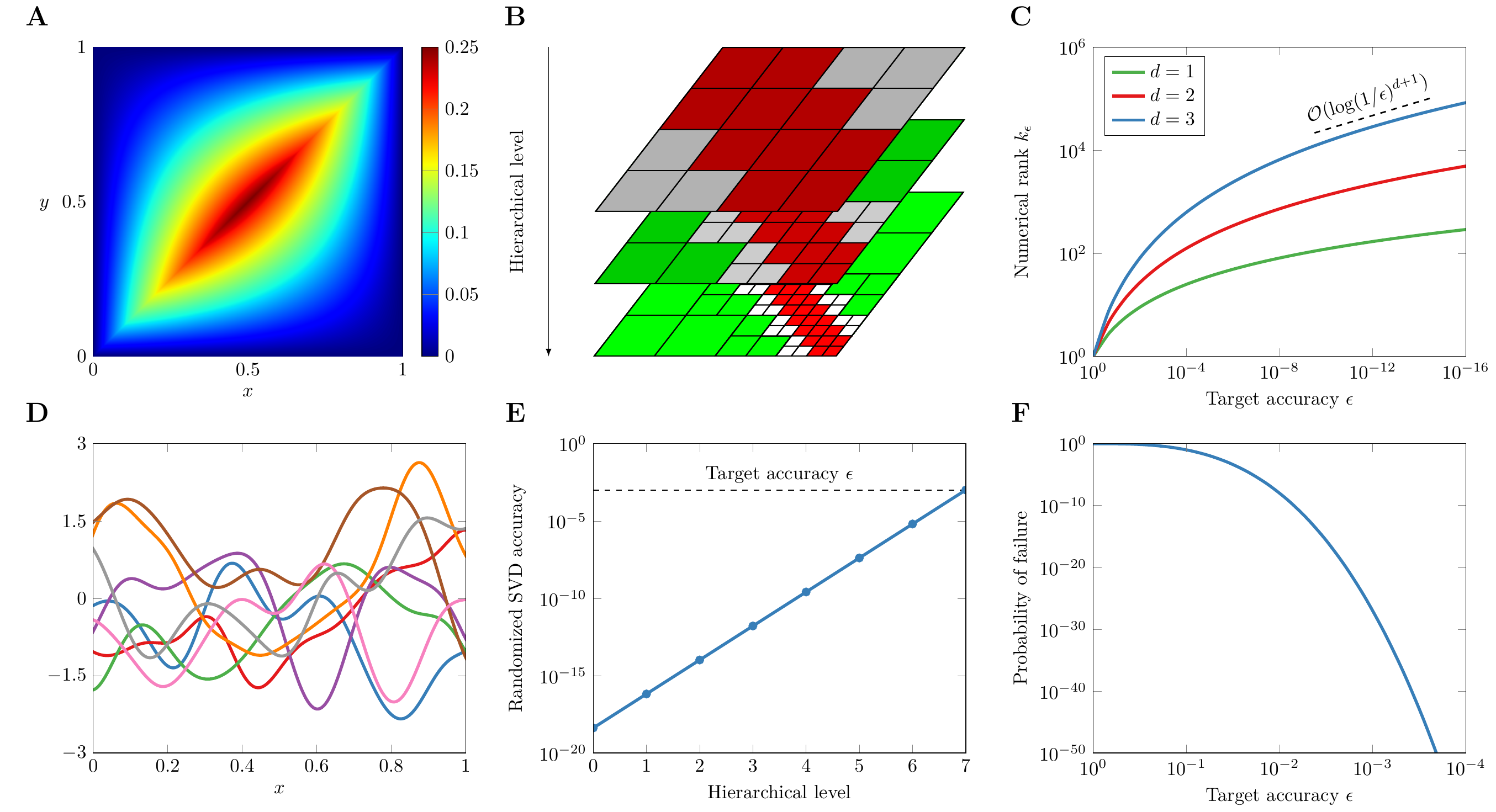}
    \end{overpic}
    \caption{Properties of elliptic PDEs can be exploited to construct a provably data-efficient algorithm for recovering solution operators. (A) The Green's function associated with the 1D Poisson equation, which is the kernel of the solution operator. (B) We use the multi-scale (hierarchical) structure of a Green's function~\cite{lin2011fast}. (C) On well-separated domains, the Green's function has rapidly decaying singular values~\cite{bebendorf2003existence}, so it is efficiently recovered by the randomized singular value decomposition (SVD)~\cite{halko2011finding}. (D) Forcing terms for the training dataset are randomly sampled from a Gaussian process. (E) The accuracy of the randomized SVD is carefully adapted on each hierarchical level to counterbalance the potential accumulation of errors in the reconstruction process. (F) An upper bound on the probability of failure of the reconstruction algorithm as a function of $\epsilon$.}
    \label{fig_schematic}
\end{figure*}

There is a lack of understanding for the efficiency of PDE learning methods with limited training data~\cite{lu2021learning}. This work provides theoretical insights by constructing a provably data-efficient algorithm, showing that one can achieve exponential convergence when learning solution operators of elliptic PDEs.

Consider an unknown uniformly elliptic PDE in three dimensions, defined on a bounded domain $\Omega \subset\R^3$ with Lipschitz smooth boundary, with variable coefficients of the form:
\begin{equation} \label{eq_PDE}
    \L u = -\nabla \cdot(A(x)\nabla u) = f, \quad x\in \Omega,\quad  u|_{\partial\Omega}=0,
\end{equation}
where the coefficient matrix $A$ has bounded coefficient functions and is symmetric positive definite for all $x\in \Omega$. The weak assumptions on $\Omega$ and $A(x)$ allow for corner singularities and low regularity of the coefficients. The training data consists of pairs of random forcing terms $f_1,\ldots,f_N$ and corresponding solutions $u_1,\ldots,u_N$ such that $\L u_j = f_j$ for $1\leq j\leq N$. Deep learning techniques use this data to predict solutions to \cref{eq_PDE} at new forcing terms by recovering the action of the solution operator $\F$, which is given by
\begin{equation} \label{eq_integral}
    \F(f) = \int_{\Omega} G(x,y) f(y) \d y,
\end{equation}
where $G$ is the associated Green's function. For example, we visualize in~\cref{fig_schematic}A the Green's function associated with the 1D Poisson equation. The random forcing terms in the training dataset are sampled from a Gaussian process (GP), i.e., they follow a multivariate Gaussian distribution when sampled on a grid, and the covariance kernel determines the correlation between the function's entries and its smoothness.

Recent work~\cite{boulle2022learning} proves that for any $\epsilon>0$ and 3D elliptic PDEs, a large number of input-output training pairs of size about $\mathcal{O}(\epsilon^{-6})$ is sufficient to recover an $\epsilon$-approximation $\tilde{\F}$ to $\F$ such that
\[
    \|\F - \tilde{\F}\|_{2} \leq \epsilon \|\F\|_{\HS},
\]
where $\|\cdot\|_2$ is the solution operator norm and $\|\cdot\|_{\text{HS}}$ is the Hilbert--Schmidt norm.  Once the $\epsilon$-approximation to $\F$ has been constructed, $\tilde{\F}$ can be used to study the stability and regularity of solutions of the PDE. For example, to see whether small perturbations of the input function lead to small changes in the output solution, or whether the solution has certain smoothness or decay properties for all forcing terms. Moreover, $\tilde{\F}$ can be used in numerical methods for approximating the solution of the PDE. By discretizing the input function and applying $\tilde{\F}$ as a surrogate for $\F$, one can obtain a numerical solution of the PDE that approximates the true solution. The integral kernel associated with the Hilbert--Schmidt operator $\tilde{\F}$ is also of interest, as it is an approximation to the Green's function, which can be exploited to recover linear conservation laws, symmetries, boundary effects, and dominant modes~\cite{boulle2022data}.

Our main result dramatically improves the required amount of training data to construct an $\epsilon$-approximation to $\F$ by exploiting the hierarchical structure of $G$~\cite{bebendorf2003existence} and randomized linear algebra techniques~\cite{halko2011finding,martinsson2020randomized}. We derive a randomized algorithm that provably succeeds with exceptionally high probability and needs a training dataset size of only $\mathcal{O}(\log(1/\epsilon)^{5}[\log(\log(1/\epsilon))+\log(1/\Gamma_\epsilon)]^{4})$ input-output pairs.

\begin{theorem} \label{th_learning_rate}
    Let $\epsilon>0$ be sufficiently small, and $\F$ be the solution operator associated with a 3D uniformly elliptic PDE of the form in~\cref{eq_PDE}. There exists a randomized
    algorithm that constructs an $\epsilon$-approximation $\tilde{\F}$ to $\F$ such that
    \[
        \|\F - \tilde{\F}\|_2 \leq \epsilon\|\F\|_{\HS},
    \]
    using $\mathcal{O}(\log(1/\epsilon)^{5}[\log(\log(1/\epsilon))+\log(1/\Gamma_\epsilon)]^{4})$ input-output pairs with probability $\geq 1-e^{-\log(1/\epsilon)^3}$.
\end{theorem}

The main contribution of \cref{th_learning_rate} is a theoretical upper bound on the amount of training data required in elliptic PDE learning problems, which should deepen our understanding of existing deep learning techniques. Hence, the exponential convergence rate in \cref{th_learning_rate} matches the one observed in the deep learning experiments of \cref{fig_exp}A. We believe that this learning rate is near-optimal, as it exploits the multi-scale structure of Green's functions (see~\cref{fig_schematic}B,C) and depends on the training dataset.  The factor $0<\Gamma_\epsilon\leq 1$ measures the quality of the training dataset at probing the dominant modes of the PDE, and a technical definition is available in the SI Appendix. We emphasize that the error bound must include a factor that quantifies the quality of the training dataset. If the forcing terms are too smooth, then $\Gamma_\epsilon$ is small. In contrast, choosing the covariance kernel of the GP such that the sampled functions are oscillatory usually ensures that $\Gamma_\epsilon$ is reasonable for learning $G$. In short, a small number of sufficiently diverse forcing terms is required (see~\cref{fig_schematic}D).

The algorithm constructed in the proof of~\cref{th_learning_rate} achieves an approximation error measured in the solution operator norm. This mimics the typical measurement of accuracy of PDE learning techniques by comparing true and predicted solutions on a testing dataset of square-integrable forcing terms. Additionally, \cref{th_learning_rate} employs random input-output pairs, where the forcing terms are sampled from a GP,  so there is always some probability of failure. Fortunately, we show this probability is exceptionally small. For $\epsilon<10^{-3}$, failure is a once-in-a-cosmic-epoch event (see~\cref{fig_schematic}F).

\cref{th_learning_rate} is  challenging  to prove, and the whole argument is in the SI Appendix. The proof relies on the fact that the solution operator associated with a 3D elliptic PDE is an integral operator in the form of~\cref{eq_integral}. Firstly, the Green's functions related to 3D elliptic operators are square-integrable and have a bounded decay rate away from the diagonal of $\Omega\times\Omega$~\cite{gruter1982green}. Secondly, they possess a hierarchical structure~\cite{bebendorf2003existence} in the sense that they have rapidly decaying singular values when restricted to off-diagonal parts of the domain (green blocks in \cref{fig_schematic}B). We leverage the hierarchical structure, which has been historically exploited by fast solvers, in a data-driven context where the PDE is unknown. Combining these properties enables a generalization of the randomized SVD~\cite{halko2011finding} known as the peeling algorithm~\cite{lin2011fast} to simultaneously learn the off-diagonal blocks at any level of the hierarchy.

While the peeling algorithm is traditionally used to recover hierarchical matrices efficiently from matrix-vector products, we generalize it to approximate infinite-dimensional integral operators. To do so, we leverage insights from recent work that extends the peeling algorithm to arbitrary hierarchical partitions and dimensions~\cite{levitt2022randomized}. This gives us a strategy to recover the Green's function level-by-level. However, proving the stability of peeling is an open question in numerical linear algebra. This is because the approximation errors from one level can potentially accumulate exponentially at later levels, thus degrading the convergence rate~\cite{boulle2022learning,lin2011fast}.

We overcome this theoretical obstacle in the infinite-dimensional context by requiring an adaptive approximation accuracy at each level of the hierarchy. The peeling algorithm ensures that the large-scale features of a Green's function are first learned to high accuracy by the randomized SVD. Then, we progressively decrease the accuracy requirement at subsequent levels, ensuring an overall $\epsilon$-approximation on each level of the partition at the end (see~\cref{fig_schematic}E). The rapidly decaying singular values of the Green's function on off-diagonal parts of the domain (see~\cref{fig_schematic}C) enable us to maintain a near-optimal exponential convergence rate with respect to the size of the training dataset. We then construct a global $\epsilon$-approximant by neglecting $G$ near the diagonal of the domain.

As one usually employs deep learning techniques to learn solution operators, our theoretical contributions  can also lead to practical benefits. We believe that future training datasets benefit from taking into account prior knowledge of the PDE to improve the quality of the forcing terms at learning the solution operator. Similar ideas have already been employed in the field of visual object recognition through data-augmentation techniques. There is also an opportunity to design neural network architectures with hierarchical structures to capture the long-range interactions in PDE models. Finally, enforcing a different accuracy at different scales might improve the computational efficiency of existing PDE learning approaches.

In summary, we constructed a randomized algorithm that provably achieves an exponential convergence rate for approximating the solution operator associated with 3D elliptic PDEs in terms of the size of the training dataset. This provides a theoretical explanation for the observed performance of recent deep learning techniques in PDE learning. The proof techniques can be adapted to include elliptic PDEs in any dimension and time-dependent PDEs~\cite{boulle2022parabolic}. Recovering solution operators associated with hyperbolic PDEs, like wave equations, remains a significant open challenge. Moving forward, we plan to ramp up PDE learning techniques to handle noisy experimental data, deal with data from emerging transient dynamics, and enforce conservation laws onto our solutions.

\subsection*{Data availability}
All data and codes used in this article are publicly available on GitHub at \url{https://github.com/NBoulle/pde-learning}. The proof of \cref{th_learning_rate} and details of the numerical experiments are available in the SI Appendix.

\acknow{This work is supported by National Science Foundation grants DMS-1952757, DMS-2045646, and DGE-2139899. N.B. was supported by an INI-Simons Postdoctoral Research Fellowship.}

\showacknow{} 

\bibliography{pnas-sample}

\begin{thebibliography}{10}

\bibitem{karniadakis2021physics}
GE Karniadakis, et~al., Physics-informed machine learning.
\newblock {\em\protect\JournalTitle{Nat. Rev. Phys.}} \textbf{3}, 422--440
  (2021).

\bibitem{lecun2015deep}
Y LeCun, Y Bengio, G Hinton, Deep learning.
\newblock {\em\protect\JournalTitle{Nature}} \textbf{521}, 436--444 (2015).

\bibitem{raissi2020hidden}
M Raissi, A Yazdani, GE Karniadakis, Hidden fluid mechanics: Learning velocity
  and pressure fields from flow visualizations.
\newblock {\em\protect\JournalTitle{Science}} \textbf{367}, 1026--1030 (2020).

\bibitem{lu2021learning}
L Lu, P Jin, G Pang, Z Zhang, GE Karniadakis, {Learning nonlinear operators via
  DeepONet based on the universal approximation theorem of operators}.
\newblock {\em\protect\JournalTitle{Nat. Mach. Intell.}} \textbf{3}, 218--229
  (2021).

\bibitem{li2020fourier}
Z Li, et~al., {Fourier Neural Operator for Parametric Partial Differential
  Equations} in {\em ICLR}.
\newblock (2021).

\bibitem{boulle2022data}
N Boull{\'e}, CJ Earls, A Townsend, {Data-driven discovery of Green’s
  functions with human-understandable deep learning}.
\newblock {\em\protect\JournalTitle{Sci. Rep.}} \textbf{12}, 1--9 (2022).

\bibitem{boulle2020rational}
N Boull{\'e}, Y Nakatsukasa, A Townsend, Rational neural networks in {\em
  NeurIPS}.
\newblock Vol.{}~33, pp. 14243--14253 (2020).

\bibitem{lin2011fast}
L Lin, J Lu, L Ying, Fast construction of hierarchical matrix representation
  from matrix-vector multiplication.
\newblock {\em\protect\JournalTitle{J. Comput. Phys.}} \textbf{230}, 4071--4087
  (2011).

\bibitem{bebendorf2003existence}
M Bebendorf, W Hackbusch, {Existence of $\mathcal{H}$-matrix approximants to
  the inverse FE-matrix of elliptic operators with $L^\infty$-coefficients}.
\newblock {\em\protect\JournalTitle{Numer. Math.}} \textbf{95}, 1--28 (2003).

\bibitem{halko2011finding}
N Halko, PG Martinsson, JA Tropp, {Finding structure with randomness:
  Probabilistic algorithms for constructing approximate matrix decompositions}.
\newblock {\em\protect\JournalTitle{SIAM Rev.}} \textbf{53}, 217--288 (2011).

\bibitem{boulle2022learning}
N Boull{\'e}, A Townsend, Learning elliptic partial differential equations with
  randomized linear algebra.
\newblock {\em\protect\JournalTitle{Found. Comput. Math.}} pp. 1--31 (2022).

\bibitem{martinsson2020randomized}
PG Martinsson, JA Tropp, {Randomized numerical linear algebra: Foundations and
  algorithms}.
\newblock {\em\protect\JournalTitle{Acta Numer.}} \textbf{29}, 403--572 (2020).

\bibitem{gruter1982green}
M Gr{\"u}ter, KO Widman, The {G}reen function for uniformly elliptic equations.
\newblock {\em\protect\JournalTitle{Manuscripta Math.}} \textbf{37}, 303--342
  (1982).

\bibitem{levitt2022randomized}
J Levitt, PG Martinsson, Randomized compression of rank-structured matrices
  accelerated with graph coloring.
\newblock {\em\protect\JournalTitle{arXiv preprint arXiv:2205.03406}} (2022).

\bibitem{boulle2022parabolic}
N Boull{\'e}, S Kim, T Shi, A Townsend, {Learning Green’s functions
  associated with time-dependent partial differential equations}.
\newblock {\em\protect\JournalTitle{J. Mach. Learn. Res.}} \textbf{23}, 1--34
  (2022).

\end{thebibliography}



\section*{Supplementary material}

**Supporting Information Text**

## 1. Theoretical results on Green's functions

We consider a second-order uniformly elliptic partial differential operator $\mathcal{L} : \mathcal{H}^2(\Omega) \cap \mathcal{H}_0^1(\Omega) \to L^2(\Omega)$ on a bounded domain $\Omega \subset \mathbb{R}^d$ in spatial dimension $d=3$ with a Lipschitz smooth boundary (see (1, Def. 6.2.33)). Here, $\mathcal{H}^k(\Omega)$ denotes the $k$th Sobolev space, $H_0^k(\Omega)$ is the closure of the infinitely differentiable functions compactly supported in $\Omega$ in $H^k(\Omega)$, and $L^2(\Omega)$ is the space of square-integrable functions over $\Omega$. We assume that the operator $\mathcal{L}$ takes the following divergence form:

$$\mathcal{L}u = -\nabla \cdot (A(x)\nabla u), \quad x \in \Omega, \quad u|_{\partial\Omega} = 0. \qquad [1]$$

Here, the coefficient matrix $A(x) \in \mathbb{R}^{d \times d}$ is a symmetric positive definite matrix for every $x \in \Omega$ with $\kappa_C = \sup\{\lambda_{\max}(x)/\lambda_{\min}(x) \mid x \in \Omega\} < \infty$, and $A$ has bounded coefficient functions, i.e., $A_{ij} \in L^\infty(\Omega)$ for $1 \leq i,j \leq d$. Also, $\lambda_{\min}(x)$ and $\lambda_{\max}(x)$ denote the smallest and largest eigenvalues of $A(x)$. Under these conditions, it is known (2) that there is a Green's function $G : \Omega \times \Omega \to \mathbb{R} \cup \{\infty\}$ associated with Eq. (1) so that the solution operator associated with $\mathcal{L}$ can be written as

$$\mathscr{F}(\phi)[x] = \int_\Omega G(x,y)f(y)\,\mathrm{d}y, \quad f \in L^2(\Omega). \qquad [2]$$

In three dimensions, $G$ is square integrable (2), and its associated integral operator is a Hilbert–Schmidt (HS) operator (3).

PDE learning techniques aim to approximate the solution operator associated with an unknown PDE, such as Eq. (1), from pairs of input-output functions $\{(f_j, u_j)\}_{j=1}^N$ (4). The forcing terms $f_j$ are usually sampled from a Gaussian process (GP) with a user-prescribed covariance kernel (5–7), and the associated solutions $u_j$ are acquired by solving the equation $\mathcal{L}u_j = f_j$ using a black-box solver (i.e., through direct numerical simulations or by performing a physical experiment). In the past few years, several approaches have employed neural networks to approximate the solution operator in a wide range of problems (5–11). While these methods have been very successful in practice, theoretical results are limited and mostly focus on the type and complexity of neural network architectures needed to approximate a given solution operator (7, 10, 12).

In this work, we focus on characterizing the sample complexity associated with solution operators of elliptic problems in three dimensions to understand the amount of training data needed to reach a given tolerance $0 < \epsilon < 1$. A number of studies have provided algebraic upper bounds of the form of $N = \mathcal{O}(\epsilon^{-2d})$ on the sample complexity for elliptic problems in three dimensions (13–15). This bound naturally extends to time-dependent (parabolic) problems by changing the norm in which the error is measured to an $L^1$-norm to obtain a sample complexity of $\mathcal{O}(\epsilon^{-(3+d)/2}\log(1/\epsilon))$ (16). However, the properties of Green's functions, such as their low-rank hierarchical structures (16, 17), and deep learning experiments (6, 7) suggest an exponentially smaller (poly-logarithmic) sample complexity of $\mathcal{O}(\log(1/\epsilon)^\alpha)$, for some power $\alpha \geq 1$. Recent work shows that such bounds can be attained by employing a recovery algorithm based on a sparse Cholesky factorization (18, 19). However, the proof technique requires queries of the solution operator at deterministic piecewise polynomial inputs, which is not close to the setting employed by state-of-the-art deep learning approaches. A key difficulty in improving the polynomial error bound from (13) to a poly-logarithmic sample complexity comes from the hierarchical structure of the Green's function. Hence, while one can exploit the hierarchical structure to greatly reduce the number of training pairs using a recursive peeling algorithm (20–23), approximation errors may accumulate exponentially during the procedure and significantly deteriorate the error bound. Additionally, designing a provably stable algorithm for recovering $n \times n$ hierarchical matrices from $\mathcal{O}(\log n)$ matrix-vector products is an open problem in numerical linear algebra (13, Sec. 5.1).

In this paper, we generalize the peeling algorithm for hierarchical matrices to infinite dimensions and control the potential accumulation of errors by adaptively refining the tolerance. Our main theorem (see Theorem 2) provides a poly-logarithmic upper bound for the sample complexity of solution operators associated with elliptic problems in three dimensions. We note that one has to express the error in the solution operator norm $\|\cdot\|$, defined in Definition 1. This metric is consistent with the usual way of measuring the model error on a test set in deep learning approaches (5–11).

**Definition 1** (Operator and HS norms). *Let $\mathscr{F} : L^2(\Omega) \to L^2(\Omega)$ be the solution operator associated with $\mathcal{L}$ and Green's function kernel $G$. Its operator norm $\|\cdot\|_2$ and HS-norm $\|\cdot\|_{\mathrm{HS}}$ are defined as (3, Chapt. 3)*

$$\|\mathscr{F}\|_2 = \sup\{\|\mathscr{F}(f)\|_{L^2(\Omega)} \mid f \in L^2(\Omega), \|f\|_{L^2(\Omega)} = 1\}, \qquad \|\mathscr{F}\|_{\mathrm{HS}} = \|G\|_{L^2(\Omega \times \Omega)}.$$

The operator norm is a generalization of the spectral norm $\|\cdot\|_2$ for matrices to Hilbert–Schmidt operators and could alternatively be defined as the largest singular value of $\mathscr{F}$, while the HS-norm is a generalization of the Frobenius norm.

**Theorem 2.** *Let $0 < \epsilon < 1$ be sufficiently small, $\Omega \subset \mathbb{R}^d$ be a bounded Lipschitz domain in dimension $d=3$, and $\mathcal{L}$ be an elliptic partial differential operator in the form given by Eq. (1). If $\mathscr{F} : L^2(\Omega) \to L^2(\Omega)$ is the solution operator associated with $\mathcal{L}$, then there is a randomized algorithm that constructs an approximation $\tilde{\mathscr{F}}$ of $\mathscr{F}$ using $N = \mathcal{O}(\log(1/\epsilon)^{d+2}[\log(\log(1/\epsilon)) + \log(1/\Gamma_\epsilon)]^{d+1})$ input-output pairs $\{(f_j, u_j)\}_{j=1}^N$, such that*

$$\|\mathscr{F} - \tilde{\mathscr{F}}\|_2 \leq \epsilon \|\mathscr{F}\|_{\mathrm{HS}},$$

*with probability greater than $1 - e^{-\log(1/\epsilon)^d}$. The factor $0 < \Gamma_\epsilon \leq 1$ is a measure of the quality of the forcing terms at approximating the eigenfunctions of $\mathcal{L}$ (see Eq. (25)).*

**Nicolas Boullé, Diana Halikias and Alex Townsend**

The proof of Theorem 2 is summarized in Algorithm 1 and occupies the rest of this Supplementary Information Text. It exploits two theoretical properties of Green's functions: (1) the low-rank structure on well-separated domains (see Section 1A) and (2) their decay away from the singularity along the diagonal of the domain (see Section 1B). In Section 2, we review existing results for the discrete analogue of Green's function recovery from input-output pairs, which is equivalent to hierarchical matrix recovery from matrix-vector products. We describe the randomized singular value decomposition (see Section 2A) and the peeling algorithm in Section 2C for reconstructing hierarchical low-rank matrices from matrix-vector products (20, 24–26). We also include a perturbation analysis of the randomized singular value decomposition under additive perturbation errors in Section 2B. In Section 3, we extend these results to the continuous case of Green's function recovery. We describe the previously established randomized singular value decomposition for Hilbert–Schmidt operators (see Section 3A), then use this result to develop an infinite-dimensional analogue of the peeling algorithm (see Section 3B). These ideas are finally combined in Sections 3C to 3E to show that one can stably recover Green's functions associated with elliptic operators of the form of Eq. (1) using a poly-logarithmic number of input-output functions (*i.e.*, a spectrally efficient learning rate).

---

**Algorithm 1** Learning the solution operator via input-output functions.

---

**Input:** Black-box numerical solver associated with $\mathcal{L}$, GP covariance kernel $K$, tolerance $0 < \epsilon < 1$
**Output:** Approximation $\tilde{\mathscr{F}}$ of the solution operator $\mathscr{F}$ within relative error $\epsilon$
1: Determine the number of hierarchical levels $N_\epsilon$ using Green's functions' off-diagonal decay (Eq. (7))
2: Set initial tolerance to $\epsilon_1 = \epsilon^p \ll \epsilon$
3: **for** $L = 1 : N_\epsilon$ **do**
4:      Sample the GP with covariance kernel $K$ (Section 3B)
5:      Use peeling to sketch the Green's function on new admissible domains (Sections 2C and 3B)
6:      Approximate the Green's function on new admissible domains using the randomized SVD with tolerance $\epsilon_L$ (Section 3A)
7:      Increase the randomized SVD tolerance (Proposition 10)
8: Pad the approximant $\tilde{\mathscr{F}}$ with zeros over the remaining non-admissible domains (Section 3E)

---

**A. Low-rank structure on well-separated domains.** Green's functions of elliptic operators have low numerical rank on well-separated domains $X, Y \subset \Omega$. More precisely, we say that $X$ and $Y$ satisfy a strong admissible condition if

$$\text{dist}(X, Y) \geq \rho \max\{\text{diam}\, X, \text{diam}\, Y\}, \qquad [3]$$

where $\rho > 0$ is an arbitrary constant measuring the relative distance between $X$ and $Y$. Let $0 < \epsilon < 1$ be a target relative accuracy. Bebendorf & Hackbusch (17, Thm. 2.8) proved that when the pair $(X, Y)$ is strongly admissible, there exists a separable approximation $G_k(x, y) = \sum_{i=1}^{k} u_i(x) v_i(y)$ of the Green's function such that

$$\|G - G_k\|_{L^2(X \times Y)} \leq \epsilon \|G\|_{L^2(X \times \hat{Y})},$$

where $k \leq k_\epsilon$ and $k_\epsilon = M \log(1/\epsilon)^{d+1}$ for some constant $M > 0$ that depends on $\mathcal{L}$. Here, $Y \subset \hat{Y} \subset \Omega$ denotes a domain slightly larger than $Y$ (see (17, Thm. 2.8) for the precise definition of $\hat{Y}$). Using the Eckart–Young–Mirsky theorem (27, 28), we deduce that Green's functions associated with elliptic operators have numerical rank bounded by $k_\epsilon$ on well-separated domains. As noted in (13), this property enables the use of randomized numerical linear algebra techniques for approximating Green's functions from input-output pairs. A pair $(X, Y) \subset \Omega \times \Omega$ that does not satisfy Eq. (3) is called non-admissible.

This property leads to a hierarchical partition of $\Omega \times \Omega$ into admissible and non-admissible domains (24–26). Without loss of generality, we assume that the domain of the hierarchical partition is $\Omega = [0, 1]^d$. Otherwise, one may rescale and shift $\Omega$ such that $\Omega \subset [0, 1]^d$ and consider the intersection of the partition of $[0, 1]^d$ and $\Omega$. Since $G$ is not of low rank on the initial domain $\Omega \times \Omega$, the domain $[0, 1]^d$ is dyadically partitioned into $2^d$ smaller subdomains given by halving each of the $d$ intervals $[0, 1]$. This leads to $N^{(L)} \leq 6^d 2^{dL}$ admissible domains at the level $1 \leq L \leq N$ of the hierarchical partition, where $N \geq 1$ is the number of hierarchical levels. Here, we chose an admissibility constant $\rho = 1/\sqrt{d}$ in Eq. (3) so that non-admissible domains are neighboring boxes. We note that the rescaling procedure is equivalent to embedding $\Omega$ into a cube whose side length depends on the diameter of $\Omega$. Then, one constructs a hierarchical partition of the cube and intersects each component with $\Omega$ to obtain the desired partition. Therefore, the constant in front of the learning rate estimate in Theorem 2 depends implicitly on the geometry of $\Omega$ such as its diameter. A domain with poor aspect ratio leads to a suboptimal partition and therefore we do not expect to have the best constant in Theorem 2.

**B. Off-diagonal decay.** We determine the number of hierarchical levels using a second property of Green's functions associated with elliptic operators, known as off-diagonal decay, which controls the magnitude of Green's functions on non-admissible domains. Following (2, Thm. 1.1), we know that a Green's function in three dimensions decays away from the diagonal:

$$G(x, y) \leq \frac{c_{\kappa_C}}{|x - y|} \|G\|_{L^2(\Omega \times \Omega)}, \quad x, y \in \Omega,\ x \neq y, \qquad [4]$$

where $c_{\kappa_C}$ is a constant depending on the spectral condition number of the operator $\mathcal{L}$ defined in Eq. (1), and $|\cdot|$ denotes Euclidean distance. For a non-admissible pair $X \times Y \subset \Omega \times \Omega$ such that $\text{dist}(X, Y) < \rho \max\{\text{diam}\, X, \text{diam}\, Y\}$, the $L^2$-norm of $G$

over $X \times Y$ can be bounded by integrating Eq. (4), following the argument of (13, Sec. 4.2). Let $r = (2+\rho)\max\{\text{diam } X, \text{diam } Y\}$ be a constant depending on the size of $X$ and $Y$, then the $L^2$-norm of the Green's function associated with $\mathcal{L}$ satisfies

$$\|G\|_{L^2(X \times Y)} \leq C_\rho c_{\kappa_C} r^2 \|G\|_{L^2(\Omega \times \Omega)}, \qquad [5]$$

where $C_\rho = \sqrt{2}\pi/\sqrt{3(2+\rho)^3}$ is an integration constant.

Using the hierarchical partition of $\Omega \times \Omega$ introduced in Section 1A, we find that the maximum diameter of the non-admissible domains $X \times Y$ decays exponentially fast with the number of hierarchical levels as

$$\max\{\text{diam } X, \text{diam } Y\} \leq \frac{\sqrt{d}}{2^{N+1}}, \qquad [6]$$

since $X$ and $Y$ are contained in cubes of side length $1/2^N$. We select $\rho = 1/\sqrt{d}$ and the number of hierarchical levels, $N_\epsilon$, such that $G$ has relative $L^2$-norm bounded by $0 < \epsilon < 1$ on each non-admissible domain of the partition. Finally, combining Eqs. (5) and (6) yields

$$N_\epsilon = \frac{1}{2}\log_2(1/\epsilon) + \frac{1}{2}\log_2\left(c_{\kappa_C}\pi\sqrt{6\sqrt{2} + \sqrt{6}}\right) - 1 \sim \frac{1}{2\log 2}\log(1/\epsilon), \quad \epsilon \to 0. \qquad [7]$$

**Remark 3** (Generalization to other systems). *Elliptic operators (29) in dimensions 1 and 2 and elliptic operators with lower order terms (30) also admit a Green's function with a hierarchical structure and off-diagonal decay. The rate of off-diagonal decay in these cases results in a different constant $N_\epsilon$, so $N_\epsilon$ is dimension dependent. Thus, we suspect that the results presented in this work and and a version of Theorem 2 generalize with slightly different constants to these systems.*

## 2. Fast recovery of hierarchical matrices from matrix-vector products

A hierarchical off-diagonal low rank (HODLR) matrix $\boldsymbol{H}$ is a block-structured matrix which often arises from integral or differential equation problems (31–33), *e.g.*, by discretizing a Green's function (see Section 1A). The matrix is hierarchically partitioned into sub-blocks, where the blocks located away from the diagonal are of low rank. A pair of admissible blocks $(X, Y)$ satisfies a so-called weakly admissible condition analogous to Eq. (3) with $\rho$ if $\text{dist}(X, Y) > 0$. The number of levels in the hierarchy determines the sizes of the blocks. This section describes a "peeling" recovery algorithm (20) for approximating HODLR matrices from matrix-vector products with random test vectors using the randomized singular value decomposition (21).

**A. Randomized singular value decomposition.** We first focus on recovering a low-rank matrix using matrix-vector products with test vectors. As this is a fundamental task in numerical linear algebra, several algorithms have been proposed using different embedding techniques and various random test vectors (34), such as random perturbations (35), sparse sign matrices (36–39), and subsampled trigonometric transforms (35, 40–42). The randomized singular value decomposition (SVD) is one of the most popular algorithms for constructing a low-rank approximant of a large matrix from matrix-vector products with random Gaussian test vectors (43). While the probabilistic error analysis performed by Halko et al. applies when the input vectors are standard Gaussian, the randomized SVD has been recently analyzed for Gaussian input vectors with correlated entries determined by a general covariance matrix (13, 44). As we shall see later in Section 3A, this generalization enables the application of the randomized SVD in infinite dimensions to compute low-rank approximants of Hilbert–Schmidt integral operators and can be used to approximate off-diagonal low-rank blocks of Green's functions (see Section 1B and (13)).

Let $\boldsymbol{A}$ be an $m \times n$ real matrix, where $m \geq n$, with SVD given by $\boldsymbol{A} = \boldsymbol{U}\boldsymbol{\Sigma}\boldsymbol{V}^*$, where $\boldsymbol{U} \in \mathbb{R}^{m \times m}$ and $\boldsymbol{V} \in \mathbb{R}^{n \times n}$ have orthonormal columns, and $\boldsymbol{\Sigma}$ is an $m \times n$ rectangular diagonal matrix containing the singular values $\sigma_1(\boldsymbol{A}) \geq \cdots \geq \sigma_n(\boldsymbol{A}) \geq 0$ of $\boldsymbol{A}$. Let $k \geq 1$ be the target rank, $p \geq 2$ an oversampling parameter, and $\boldsymbol{\Omega} \in \mathbb{R}^{n \times (k+p)}$ a random test matrix with independent and identically distributed (i.i.d.) columns following a multivariate Gaussian distribution with mean 0 and covariance matrix $\boldsymbol{K} \in \mathbb{R}^{n \times n}$, *i.e.*, $\boldsymbol{\Omega}(:, j) \sim \mathcal{N}(0, \boldsymbol{K})$ for $1 \leq j \leq k+p$. It is convenient to partition the reduced SVD of $\boldsymbol{A}$ as follows:

$$\boldsymbol{A} = \begin{array}{c} {\scriptstyle k \quad n-k} \\ \left[\begin{array}{cc} \boldsymbol{U}_1 & \boldsymbol{U}_2 \end{array}\right] \end{array} \begin{array}{c} {\scriptstyle k \quad n-k} \\ \left[\begin{array}{cc} \boldsymbol{\Sigma}_1 & \\ & \boldsymbol{\Sigma}_2 \end{array}\right] \end{array} \begin{array}{c} {\scriptstyle n} \\ \left[\begin{array}{c} \boldsymbol{V}_1^* \\ \boldsymbol{V}_2^* \end{array}\right] \begin{array}{c} k \\ n-k \end{array} \end{array},$$

where the matrices $\boldsymbol{\Sigma}_1$ and $\boldsymbol{\Sigma}_2$ are diagonal and contain the first $k$ and last $n-k$ singular values of $\boldsymbol{A}$, respectively. In addition, we decompose the test vector $\boldsymbol{\Omega}$ in the basis defined by the right singular vectors of $\boldsymbol{A}$ as $\boldsymbol{\Omega}_1 = \boldsymbol{V}_1^* \boldsymbol{\Omega}$ and $\boldsymbol{\Omega}_2 = \boldsymbol{V}_2^* \boldsymbol{\Omega}$.

The randomized SVD uses the following two-stage procedure for constructing a low-rank approximant to the matrix $\boldsymbol{A}$:

1. Form the matrix product $\boldsymbol{Y} = \boldsymbol{A}\boldsymbol{\Omega}$. This step requires $k+p$ matrix-vector products with $\boldsymbol{A}$.

2. Project the matrix $\boldsymbol{A}$ onto the range of $\boldsymbol{Y}$. One achieves this by constructing a matrix $\boldsymbol{Q}$ with orthonormal columns and the same column space as that of $\boldsymbol{Y}$. Then, we project $\boldsymbol{A}$ onto the range of $\boldsymbol{Y}$ as follows: $\tilde{\boldsymbol{A}} = \boldsymbol{Q}\boldsymbol{Q}^*\boldsymbol{A} = \boldsymbol{P}_{\boldsymbol{Y}}\boldsymbol{A}$, where $\boldsymbol{P}_{\boldsymbol{Y}}$ is the orthogonal projection matrix on the range of $\boldsymbol{Y}$ defined as $\boldsymbol{P}_{\boldsymbol{Y}} = \boldsymbol{Y}\boldsymbol{Y}^\dagger$. Here, $\boldsymbol{Y}^\dagger$ is the Moore–Penrose pseudo-inverse of $\boldsymbol{Y}$. Assuming that the matrix $\boldsymbol{A} = \boldsymbol{A}^*$, this projection step also requires $k+p$ matrix-vector products with $\boldsymbol{A}$ as $\tilde{\boldsymbol{A}} = \boldsymbol{Q}(\boldsymbol{A}\boldsymbol{Q})^*$; otherwise, it requires $k+p$ matrix-vector products with $\boldsymbol{A}^*$.

The approximation error in the Frobenius norm, $\|\cdot\|_F$, between the matrix $A$ and the computed low-rank approximant $\tilde{A}$ can be characterized as follows (13, Thm. 1):

$$\|A - P_Y A\|_F \leq \sqrt{1 + t^2 s^2 \frac{3}{\gamma_k} \frac{k(k+p)}{p+1} \frac{\text{Tr}(K)}{\lambda_1}} \left(\sum_{j=k+1}^n \sigma_j^2(A)\right)^{1/2},$$

which holds with probability $\geq 1 - t^{-p} - [se^{-(s^2-1)/2}]^{k+p}$, for arbitrary numbers $s, t \geq 1$. Here, $\text{Tr}(K)$ denotes the trace of the matrix $K$, *i.e.*, the sum of its eigenvalues $\lambda_1 \geq \cdots \geq \lambda_n$. The quantity $0 \leq \gamma_k \leq 1$ measures the quality of the covariance matrix for approximating the right singular vectors of $A$ and is defined as $\gamma_k = k/(\lambda_1 \text{Tr}((V_1^* K V_1)^{-1}))$. Note that when $k = p$, one can simplify the bound by choosing $t = e$ and $s = 2$ to obtain the following error bound (16, Eq. 38):

$$\|A - P_Y A\|_F \leq \left(1 + 20\sqrt{\frac{k}{\gamma_k} \frac{\text{Tr}(K)}{\lambda_1}}\right) \left(\sum_{j=k+1}^n \sigma_j^2(A)\right)^{1/2}, \qquad [8]$$

which holds with probability $\geq 1 - 2e^{-k}$. This error bound allows us to recover a HODLR matrix $H$ with high probability from matrix-vector products by approximating it block-by-block. Later in Section 2C, we combine the randomized SVD with the peeling algorithm to greatly reduce the number of test input vectors needed by exploiting the hierarchical structure of $H$. However, employing the peeling procedure leads to a potential accumulation of errors, which can only be controlled by a careful perturbation analysis of the randomized SVD.

**B. Perturbation analysis of the randomized singular value decomposition.** In this section, we assume that matrix-vector products with $A$ introduce additive perturbation errors in the sample and projection steps of the randomized SVD, such that

$$Y_{\text{noisy}} = A\Omega + E = Y + E, \quad \text{and} \quad \tilde{A} = P_{Y_{\text{noisy}}} A + Q E_P^*. \qquad [9]$$

Here, $Y$ denotes the noiseless matrix-vector products with $A$ which we cannot access, $E \in \mathbb{R}^{m \times (k+p)}$ is the additive perturbation with $\|E\|_F \leq \epsilon$ for a given $\epsilon > 0$, and $E_P \in \mathbb{R}^{n \times (k+p)}$ is the projection error satisfying $\|E_P\|_F \leq \epsilon$. While the original theoretical results on the randomized SVD are formulated in the noise-free setting (34, 43, 45–47), a large number of studies in matrix perturbation theory have considered bounding the approximation error between the low-rank approximant and exact leading singular vectors of $A$ when the observed matrix $\hat{A}$ follows a "signal-plus-noise" model, *i.e.*, it contains noise as $\hat{A} = A + E$ (48–52). Standard bounds in the literature can be refined by introducing additional assumptions on the perturbation error matrix $E$, such as specific distribution of its entries (53–63). Recently, these results have been exploited to derive explicit perturbation errors in the context of the randomized SVD (64).

Our setting differs from previous work because we do not have access to the perturbed sketch through matrix-vector products, as error is introduced additively after sketching (see Eq. (9)). Therefore, we cannot use sketching to "learn" the perturbation error and obtain efficient error bounds with classical matrix perturbation theory results. Our aim is to analyze the error between the matrix $A$ and its approximation $\tilde{A}$ computed by the randomized SVD as

$$\|A - \tilde{A}\|_F \leq \|A - P_{Y_{\text{noisy}}} A\|_F + \|Q E_P^*\|_F \leq \|A - P_{Y_{\text{noisy}}} A\|_F + \epsilon. \qquad [10]$$

Following Eq. (10), we focus on analyzing the perturbed projection error term $\|A - P_{Y_{\text{noisy}}} A\|_F$, where $Y_{\text{noisy}} = Y + E$. This term can be directly bounded using the triangular inequality as $\|A - P_{Y_{\text{noisy}}} A\|_F \leq \|A - P_Y A\|_F + \|(P_Y - P_{Y_{\text{noisy}}}) A\|_F$, and then using orthogonal projector error analysis (65–68) to analyze the projection error $(P_Y - P_{Y_{\text{noisy}}})$. However, this approach introduces technical difficulties whenever the additive error $E$ has a non-zero component in the space spanned by the leading singular vectors of $A$. Instead, we choose to directly analyze Eq. (10) to estimate the error using a first-order expansion when $\epsilon$ is sufficiently small. The following proposition is an analogue of (43, Thm. 9.1) and provides a deterministic error bound for the randomized SVD with perturbed samples.

**Proposition 4** (Deterministic error bound). *Let $A$ be an $m \times n$ matrix with $m \geq n$ and $1 \leq k \leq n$ be a target rank, and $p \geq 2$ an oversampling parameter. Choose a test vector $\Omega \in \mathbb{R}^{n \times (k+p)}$ and construct the perturbed sample matrix as $Y_{noisy} = Y + E$, where $Y = A\Omega$ and $E \in \mathbb{R}^{m \times (k+p)}$ satisfies $\|E\|_F \leq \epsilon$. Assuming that $\Omega_1 = V_1^* \Omega$ is full rank and $\epsilon > 0$ is sufficiently small, then the approximation error satisfies*

$$\|A - P_{Y_{noisy}} A\|_F^2 \leq \|\Sigma_2\|_F^2 + \|(\Sigma_2 \Omega_2 + E_2)(\Omega_1 + \Sigma_1^{-1} E_1)^\dagger\|_F^2, \qquad [11]$$

*where $E_1 = U_1^* E$ and $E_2 = U_2^* E$.*

*Proof.* The proof of Proposition 4 closely follows the proof of (43, Thm. 9.1). We first argue that the left singular vectors of $A$ do not play any role by defining the auxiliary matrix $\tilde{A}$ and perturbed sample matrix $\tilde{Y}_{\text{noisy}}$ as

$$\tilde{A} = U^* A = \begin{bmatrix} \Sigma_1 V_1^* \\ \Sigma_2 V_2^* \end{bmatrix}, \quad \text{and} \quad \tilde{Y}_{\text{noisy}} = \tilde{A}\Omega + U^* E = \begin{bmatrix} \Sigma_1 \Omega_1 + E_1 \\ \Sigma_2 \Omega_2 + E_2 \end{bmatrix}.$$

Using the unitary invariance of the Frobenius norm and (43, Prop. 8.4), we have

$$\|A - P_{Y_{\text{noisy}}}A\|_F = \|U^*(I - P_{Y_{\text{noisy}}})U\tilde{A}\|_F = \|(I - P_{U^*Y_{\text{noisy}}})\tilde{A}\|_F = \|\tilde{A} - P_{\tilde{Y}_{\text{noisy}}}\tilde{A}\|_F.$$

It is then sufficient to show that the following inequality holds: $\|\tilde{A} - P_{\tilde{Y}_{\text{noisy}}}\tilde{A}\|_F^2 \leq \|\Sigma_2\|_F^2 + \|(\Sigma_2\Omega_2 + E_2)(\Omega_1 + \Sigma_1^{-1}E_1)^\dagger\|_F^2$. To achieve this, we first remark that, since $\Omega_1$ is full rank, it has linearly independent columns and $\Omega_1\Omega_1^*$ is invertible such that $\Omega_1^\dagger := \Omega_1^*(\Omega_1\Omega_1^*)^{-1}$ is well defined. Since $1 \leq k \leq \text{rank}(A)$, the $k$th singular value of $A$ is strictly positive and $\Sigma_1$ is invertible. Let $d_{\Omega_1}$ be the function on $k \times k$ matrices defined as $d_{\Omega_1}(X) = \det\left[(\Omega_1 + \Sigma_1^{-1}X)(\Omega_1 + \Sigma_1^{-1}X)^*\right]$ for $X \in \mathbb{R}^{k \times k}$. $d_{\Omega_1}$ is continuous and strictly positive at $X = 0$. Hence, for $\epsilon$ sufficiently small, $d_{\Omega_1}(E_1) > 0$ and $(\Omega_1 + \Sigma_1^{-1}E_1)^\dagger$ is well defined. This also implies that $\Sigma_1\Omega_1 + E_1$ has full rank and, if $W$ denotes the matrix containing the first $k$ columns of $\tilde{Y}_{\text{noisy}}$ and zero afterwards, then

$$\text{range}(W) = \text{range}\left(\begin{bmatrix} I \\ 0 \end{bmatrix} \begin{matrix} k \\ n-k \end{matrix}\right), \quad W = \begin{bmatrix} \Sigma_1\Omega_1 + E_1 \\ 0 \end{bmatrix} \begin{matrix} k \\ n-k \end{matrix}.$$

The range of $W$ spans the same subspace as the first $k$ left singular vectors of the auxiliary matrix $\tilde{A}$. We then use perturbation theory for orthogonal projectors to treat the matrix $\tilde{Y}_{\text{noisy}}$ as a perturbation of $W$, and introduce a matrix $Z$ to flatten the first $k$ rows of the auxiliary sample matrix as

$$Z = \tilde{Y}_{\text{noisy}}(\Sigma_1\Omega_1 + E_1)^\dagger = \tilde{Y}_{\text{noisy}}(\Omega_1 + \Sigma_1^{-1}E_1)^\dagger\Sigma_1^{-1} = \begin{bmatrix} I \\ F \end{bmatrix}, \quad F = (\Sigma_2\Omega_2 + E_2)(\Omega_1 + \Sigma_1^{-1}E_1)^\dagger\Sigma_1^{-1}.$$

We then follow the proof of (43, Thm. 9.1) to obtain the following inequality $\|(I - P_{\tilde{Y}_{\text{noisy}}})A\|_F^2 \leq \|\Sigma^*(I - P_Z)\Sigma\|_F \leq \|F\Sigma_1\|_F^2 + \|\Sigma_2\|_F^2$. We have $\|F\Sigma_1\|_F^2 = \|(\Sigma_2\Omega_2 + E_2)(\Omega_1 + \Sigma_1^{-1}E_1)^\dagger\|_F^2$, which shows that Eq. (11) holds. □

One can easily extract explicit upper bounds from Proposition 4 when the perturbation error is located in a subspace that is orthogonal to the first $k$ left singular vectors of $A$, contained in the matrix $U_1$. Moreover, the multiplicative factor $\Sigma_1$ in the right-hand side of Eq. (11) implies that the component of the error in the space generated by the $j$th singular vector must have magnitude bounded by the $j$th singular value of $A$. Otherwise, one cannot hope to recover a good rank $k$ approximation, as the perturbation is too large in the direction of the $j$th singular vector. To alleviate this issue, we derive a probabilistic bound for the perturbed randomized SVD by combining Proposition 4 with a first-order expansion of a perturbed pseudo-inverse matrix when the perturbation magnitude $\epsilon$ is sufficiently small.

**Theorem 5** (Probabilistic error bound). *Let $A$ be an $m \times n$ matrix with $m \geq n$, $1 \leq k \leq n$ a target rank, $p \geq 2$ an oversampling parameter, and $\Omega$ be an $n \times (k+p)$ Gaussian matrix, where each column is i.i.d. and drawn from a multivariate Gaussian distribution with mean zero and covariance matrix $K$. Assume that matrix-vector products with $A$ and $A^*$ introduce an additive error of $E$, with $\|E\|_F \leq \epsilon$ for sufficiently small $0 < \epsilon < 1$. Then, the approximation error between the matrix $A$ and its low-rank approximant $\tilde{A}$ constructed by the randomized SVD satisfies*

$$\|A - \tilde{A}\|_F \leq \sqrt{1 + t^2s^2\frac{3}{\gamma_k}\frac{k(k+p)}{p+1}\frac{\text{Tr}(K)}{\lambda_1}}\left(\sum_{j=k+1}^n \sigma_j^2(A)\right)^{1/2} + \mathcal{O}(\epsilon), \qquad [12]$$

*with probability $\geq 1 - t^{-p} - [se^{-(s^2-1)/2}]^{k+p}$.*

*Proof.* Combining Eq. (10) and Proposition 4 yields

$$\|A - \tilde{A}\|_F \leq \|A - P_{Y_{\text{noisy}}}A\|_F + \epsilon \leq \|\Sigma_2\|_F + \|(\Sigma_2\Omega_2 + E_2)(\Omega_1 + \Sigma_1^{-1}E_1)^\dagger\|_F + \epsilon. \qquad [13]$$

Let $X$ be the $k \times (k+p)$ matrix such that $\epsilon X = \Sigma_1^{-1}E_1$. By the same argument of the proof of Proposition 4, the matrix $\Omega_1 + \epsilon X$ has full rank for $\epsilon$ sufficiently small. Then, the matrix $(\Omega_1 + \epsilon X)(\Omega_1 + \epsilon X)^*$ is invertible, and we can compute a first-order expansion of the pseudo-inverse using an expansion for the matrix inverse as

$$(\Omega_1 + \epsilon X)^\dagger = (\Omega_1 + \epsilon X)^*[(\Omega_1 + \epsilon X)(\Omega_1 + \epsilon X)^*]^{-1} = (\Omega_1 + \epsilon X)^*[\Omega_1\Omega_1^* + \epsilon(X\Omega_1^* + \Omega_1 X^* + \epsilon XX^*)]^{-1} = \Omega_1^\dagger + \mathcal{O}(\epsilon).$$

Therefore, the second term in the right-hand side of Eq. (13) satisfies $\|(\Sigma_2\Omega_2 + E_2)(\Omega_1 + \Sigma_1^{-1}E_1)^\dagger\|_F = \|\Sigma_2\Omega_2\Omega_1^\dagger\|_F + \mathcal{O}(\epsilon)$, since $\|E_2\|_F \leq \|E\|_F \leq \epsilon$ and $\|E_1\|_F \leq \epsilon$. We conclude that the approximation error is bounded for $\epsilon$ sufficiently small as $\|A - \tilde{A}\|_F \leq \|\Sigma_2\|_F + \|\Sigma_2\Omega_2\Omega_1^\dagger\|_F + \mathcal{O}(\epsilon)$, which we can bound using the proof of (13, Thm. 1). □

**C. Peeling algorithm for weakly admissible hierarchical matrices.** We now consider a symmetric $n \times n$ hierarchical matrix $H$, whose off-diagonal blocks are of low rank, and hence can be approximated by the randomized SVD. The peeling algorithm (20, 69) uses a recursive elimination procedure to efficiently utilize each input-output pair to recover $H$ in $\mathcal{O}(\log n)$ matrix-vector products. The main idea is to recover low-rank blocks of $H$ level-by-level, *i.e.*, recover $H$ one level at a time, starting with the largest blocks, working towards the diagonal, and finally recovering the finest block size at a chosen stopping point.

A hierarchical partition of $H$ is obtained by halving the index set of the rows and columns at each level. Therefore, if $H$ is a symmetric rank-$k$ $n \times n$ HODLR matrix, it has $N = \lfloor \log_2(n) \rfloor$ levels, and the following block structure at the first level:

$$H = \begin{pmatrix} H_1 & H_2 \\ H_2^\top & H_4 \end{pmatrix}, \qquad [14]$$

where $H_2$ and $H_2^\top$ are considered off-diagonal blocks and are of rank at most $k$. The off-diagonal blocks are colored in green in Eq. (14) and are usually called admissible blocks as they satisfy a weakly admissible condition analogue to Eq. (3). The blocks along the diagonal, colored in red in Eq. (14), are considered non-admissible. Further hierarchical levels are obtained by recursively subdividing the first two non-admissible blocks, $H_1$ and $H_4$, to refine the matrix along the diagonal and obtain more admissible blocks with rank $k$ as follows:

$$\begin{pmatrix} H_1 & H_2 \\ H_2^\top & H_4 \end{pmatrix} \to \begin{pmatrix} \begin{matrix} H_{11} & H_{12} \\ H_{12}^\top & H_{14} \end{matrix} & H_2 \\ H_2^\top & \begin{matrix} H_{41} & H_{42} \\ H_{42}^\top & H_{44} \end{matrix} \end{pmatrix} \to \begin{pmatrix} \begin{matrix} H_{111} & H_{112} \\ H_{112}^\top & H_{114} \end{matrix} & H_{12} & & \\ H_{12}^\top & \begin{matrix} H_{141} & H_{142} \\ H_{142}^\top & H_{144} \end{matrix} & & H_2 \\ & & \begin{matrix} H_{411} & H_{412} \\ H_{412}^\top & H_{414} \end{matrix} & H_{42} \\ H_2^\top & & H_{42}^\top & \begin{matrix} H_{441} & H_{442} \\ H_{442}^\top & H_{444} \end{matrix} \end{pmatrix}. \qquad [15]$$

Again, green matrices are of rank $k$ while red matrices need to be further partitioned to reveal off-diagonal low-rank structure.

We now describe the procedure of the peeling algorithm to construct a HODLR approximant to $H$ from matrix-vector products with random test vectors using the randomized SVD. Let $H$ be a rank-$k$ symmetric HODLR matrix partitioned as in Eq. (15). Peeling first recovers the largest off-diagonal blocks of $H$, *i.e.*, $H_2$ and $H_2^\top$, which are of size $n/2 \times n/2$ at the first hierarchical level. We begin by constructing two $n \times (k+p)$ random test vectors $X_1$ and $Y_1$, with the following block structure:

$$X_1 = \begin{pmatrix} 0 \\ x_1 \end{pmatrix}, \qquad Y_1 = \begin{pmatrix} y_1 \\ 0 \end{pmatrix}.$$

Here, $x_1$ and $y_1$ are $n/2 \times (k+p)$ matrices, where $k$ is the rank of $H_2$ and $p \geq 1$ is an oversampling parameter required by the randomized SVD. Then, the products $HX_1$ and $HY_1$ allow us to access matrix-vector products with the blocks $H_2$ and $H_2^\top$, and form $H_2 x_1$ and $H_2^\top y_1$. Next, we invoke the randomized SVD to approximate the rank-$k$ block $H_2$ by sampling the columns of $x_1$ i.i.d. from a multivariate Gaussian distribution, and setting $y_1$ to be the orthogonalized output $H_2 y_1$. Once the blocks $H_2$ and $H_2^\top$ have been recovered, we define the "level truncated" matrix $H^{(1)}$ as the matrix given by zeroing out all of the blocks of $H$ with hierarchical level greater than one:

$$H^{(1)} = \begin{pmatrix} 0 & H_2 \\ H_2^\top & 0 \end{pmatrix}.$$

We note that the blocks $H_2$ and $H_2^\top$ have been recovered exactly by the randomized SVD with probability one since we assumed that they are of rank exactly $k$.

We now generalize the procedure recursively and assume that all the sub-blocks up to the hierarchical level $1 \leq L \leq N-1$ have been approximated by the randomized SVD and peeling. The $L$-truncated matrix $H^{(L)}$ is equal to $H$ for the admissible blocks at level $L$, and zero elsewhere. At the hierarchical level $L+1$, we generate the analogous two random input matrices, $X_{L+1}$ and $Y_{L+1}$, with $(k+p)$ columns and alternating random Gaussian and zero structure. We then form the products

$$\left(H - \sum_{\ell=1}^{L} H^{(\ell)}\right) X_{L+1}, \quad \left(H - \sum_{\ell=1}^{L} H^{(\ell)}\right) Y_{L+1} \qquad [16]$$

to isolate the actions of the admissible blocks at level $L+1$, and approximate them with the randomized SVD. This process continues until the final hierarchical level $N$ is reached. Applying peeling with the randomized SVD requires a total of $2(k+p)N$ matrix-vector products to recover all the admissible blocks of $H$.

**Remark 6.** *The peeling technique described in this section easily generalizes to a non-symmetric HODLR matrix $H$ by requiring two additional matrix-vector products with $H^\top$ at each hierarchical level. This is because the randomized SVD requires matrix-vector products with the transpose of each admissible block, which doubles the total number of matrix-vector products to recover $H$ to at most $4(k+p)N$.*

As it is known in the literature (20), the peeling algorithm is not stable and can potentially introduce errors that exponentially accumulate as one goes down the levels. We characterize the magnitude of these accumulating errors for a weakly admissible HODLR matrix. However, the analysis generalizes easily to strongly admissible hierarchical matrices (see Section 2D).

**Proposition 7** (Perturbation error from sketch). *Let $1 \leq L \leq N$ be a hierarchical level and assume that each of the off-diagonal blocks of $H$ at lower levels $1 \leq \ell \leq L-1$ has been approximated to within an absolute accuracy of $0 < \tilde{\epsilon}_\ell < 1$. Let $H_i \in \mathbb{R}^{n/2^L \times n/2^L}$ be an admissible block at level $L$ and $x_i \in \mathbb{R}^{n/2^L \times k}$ an input matrix. Consider the exact sketch $Z_L = H_i x_i$, and the perturbed sketch $\tilde{Z}_L$ obtained by peeling from the input[*] $X_L = [0; x_1; \cdots; 0; x_i; 0; \cdots; x_{2^{L-1}}] \in \mathbb{R}^{n \times k}$, where the Euclidean norm of each column of the matrices $x_i$ is bounded by a constant $C > 0$. Then, the approximation error satisfies*

$$\|Z_L - \tilde{Z}_L\|_F \leq C\sqrt{k} \sum_{\ell=1}^{L-1} 2^{\frac{L-1-\ell}{2}} \tilde{\epsilon}_\ell.$$

*Proof.* It suffices to prove this result for the upper left-most off-diagonal block at level $L$, which we denote as $H_{1^{(L-1)}2}$, where $1^{(L-1)}$ represents $L-1$ concatenated 1's. This is because the perturbation argument for one block applies symmetrically to any other admissible block. Let $X_L = [0; x_1; 0; \cdots; x_{2^{L-1}}] \in \mathbb{R}^{n \times k}$ be the input matrix, such that $\|x_j\|_F \leq C\sqrt{k}$ for $1 \leq j \leq 2^{L-1}$. We consider the exact sketch $Z_L = H_{1^{(L-1)}2} x_1$ of the block $H_{1^{(L-1)}2}$ with $x_1$. Then, following Eq. (16), the perturbed sketch obtained by peeling satisfies:

$$\tilde{Z}_L = Z_L + \begin{pmatrix} I_L & 0 \end{pmatrix} (H_{1^{(L-2)}2} - \tilde{H}_{1^{(L-2)}2})[0; x_2] + \cdots + \begin{pmatrix} I_L & 0 \end{pmatrix} (H_2 - \tilde{H}_2)[0; x_{2^{L-2}+1}; \cdots; 0; x_{2^{L-1}}],$$

where $I_L$ is the $n/2^L \times n/2^L$ identity matrix, *i.e.*, $I_L$'s dimensions are the block size at level $L$. Moreover, the approximant $\tilde{H}_{1^{(\ell)}2}$ of the admissible block $H_{1^{(\ell)}2}$ at level $1 \leq \ell \leq L-1$ satisfies $\|H_{1^{(\ell)}2} - \tilde{H}_{1^{(\ell)}2}\|_F \leq \tilde{\epsilon}_\ell$. Then, the sketch error is bounded by

$$\|Z_L - \tilde{Z}_L\|_F \leq \sum_{\ell=1}^{L-1} \tilde{\epsilon}_i \sqrt{\sum_{j=1}^{2^{L-1-\ell}} \|x_{2^{L-1-\ell}+j}\|_F^2} \leq C\sqrt{k} \sum_{\ell=1}^{L-1} 2^{\frac{L-1-\ell}{2}} \tilde{\epsilon}_\ell.$$

We note that the sketch error is zero at the first hierarchical level. □

Proposition 7 also characterizes the perturbation error for the projection step of the randomized SVD in the peeling algorithm. Then, if we consider the exact sketch $Z_L^P = H_i y_i$ of an off-diagonal block $H_i$ at level $L$, and the perturbed sketch $\tilde{Z}_L^P$ obtained by peeling from the input $Y_L = [y_1; 0; \cdots; 0; y_i; 0; \cdots; 0] \in \mathbb{R}^{n \times k}$, whose components $y_1, \ldots, y_{2^{L-1}}$ have orthonormal columns. The approximation error satisfies

$$\|Z_L^P - \tilde{Z}_L^P\|_F \leq \sqrt{k} \sum_{\ell=1}^{L-1} 2^{\frac{L-1-\ell}{2}} \tilde{\epsilon}_\ell.$$

**D. Peeling algorithm for strongly admissible partitions.** We now generalize the peeling algorithm introduced in Section 2C to hierarchical matrices with arbitrary structure. As we saw in Section 1A, Green's functions associated with elliptic operators satisfy a strong admissibility condition, which implies that sub-blocks intersecting the diagonal are no longer considered admissible and may not have low numerical rank. As an example, for a symmetric block matrix $H$ in Eq. (14), the blocks $H_2$ and $H_2^\top$ are not admissible, and we instead recursively partition $H$ as follows,

$$\begin{pmatrix} H_1 & H_2 \\ H_2^\top & H_4 \end{pmatrix} \rightarrow \begin{pmatrix} H_{11} & H_{12} & H_{21} & H_{22} \\ H_{12}^\top & H_{14} & H_{23} & H_{24} \\ H_{21}^\top & H_{23}^\top & H_{41} & H_{42} \\ H_{22}^\top & H_{24}^\top & H_{42}^\top & H_{44} \end{pmatrix} \rightarrow \begin{pmatrix} \begin{matrix} H_{111} & H_{112} \\ H_{112}^\top & H_{114} \end{matrix} & \begin{matrix} H_{121} & H_{122} \\ H_{123} & H_{124} \end{matrix} & H_{21} & H_{22} \\ \begin{matrix} H_{121}^\top & H_{123}^\top \\ H_{122}^\top & H_{124}^\top \end{matrix} & \begin{matrix} H_{141} & H_{142} \\ H_{142}^\top & H_{144} \end{matrix} & \begin{matrix} H_{231} & H_{232} \\ H_{233} & H_{234} \end{matrix} & H_{24} \\ H_{21}^\top & \begin{matrix} H_{231}^\top & H_{233}^\top \\ H_{232}^\top & H_{234}^\top \end{matrix} & \begin{matrix} H_{411} & H_{412} \\ H_{412}^\top & H_{414} \end{matrix} & \begin{matrix} H_{421} & H_{422} \\ H_{423} & H_{424} \end{matrix} \\ H_{22}^\top & H_{24}^\top & \begin{matrix} H_{421}^\top & H_{423}^\top \\ H_{422}^\top & H_{424}^\top \end{matrix} & \begin{matrix} H_{441} & H_{442} \\ H_{442}^\top & H_{444} \end{matrix} \end{pmatrix}.$$

[17]

Hierarchical matrices partitioned as in Eq. (17) satisfy a strong admissible condition (see Eq. (3)) in spatial dimension $d = 1$. Partitions in higher dimensions work similarly but are harder to visualize and possess a larger bandwidth of non-admissible blocks at each level. Constructing the test vectors $X_L$ and $Y_L$, where $1 \leq L \leq N$, is crucial to reducing the number of matrix-vector products in the randomized SVD when recovering $H$. If one naively employs different test vectors at each admissible block, then the number of matrix-vector products grows exponentially with the number of levels rather than linearly.

Fortunately, peeling has been recently generalized to hierarchical matrices with arbitrary structure (23). The nature of the hierarchical partition (weakly or strongly admissible) determines the structure and number of alternating inputs in the test

---

[*]This notation denotes a matrix whose components $0, x_1, 0, \ldots, 0, x_{2^{L-1}} \in \mathbb{R}^{n/2^L \times k}$ are stacked vertically.

matrices at each hierarchical level to maintain the linear growth of the number of matrix-vector products needed with respect to the number of levels. The peeling algorithm recovers a matrix with a strongly admissible partition in exactly the same way as for weakly admissible, except the input vectors have slightly different structures to isolate the actions of the low-rank sub-blocks. This structure is derived by the generalized coloring algorithm technique in (23), which we now describe.

The peeling algorithm aims to sketch all the admissible blocks of a given level $1 \leq L \leq N$ of a hierarchical matrix $\boldsymbol{H}$. We generate a set of test vectors such that for each admissible block at level $L$, a test vector satisfies a set of sketching constraints associated with that block. More specifically, for a given admissible block, we require a test vector that samples the block and avoids contribution from non-admissible blocks and admissible blocks of the same level. As an example, if we want to sketch the block $\boldsymbol{H}_{231}^\top$ in Eq. (17), the test vector must vanish for the blocks $\boldsymbol{H}_{233}^\top$, $\boldsymbol{H}_{411}$, $\boldsymbol{H}_{412}$, $\boldsymbol{H}_{421}$, and $\boldsymbol{H}_{422}$. Of course, one could use as many test vectors as admissible blocks, but then the number of inputs would grow exponentially with the number of levels. To resolve this issue, (23) constructs inputs that satisfy the constraints for several different admissible blocks simultaneously by defining a constraint incompatibility graph. The vertices of the graph correspond to the constraint set for a particular admissible block, and vertices are connected if their constraints conflict with one another. A vertex coloring algorithm finds the minimal number of colors (chromatic number) denoted as $\chi(\boldsymbol{H}_L) \in \mathbb{N}$ for all the constraint sets at level $L$. For each color $1 \leq j \leq \chi(\boldsymbol{H}_L)$, there is a test vector $\boldsymbol{\Omega}_L^{(j)}$ that satisfies the constraints of all the vertices colored by $j$. A key insight is that the chromatic number is bounded by a constant which is independent of the level of the partition and size of the matrix (23), meaning that one can use a constant number of input matrices at each level. In particular, if $\boldsymbol{H}$ satisfies a strongly admissible condition in dimension $d \geq 1$, we have the bound $\chi(\boldsymbol{H}_L) \leq 6^d$ for each level $L$ (23). As a comparison, the chromatic number for the HODLR matrix with a weakly admissible partition analyzed in Section 2C is two.

We illustrate the input matrices for the strongly hierarchical matrix in Eq. (17), whose chromatic number is bounded above as $\chi(\boldsymbol{H}_L) \leq 6$. At the first level, *i.e.*, the second matrix in Eq. (17), there are four test vectors of the form:

$$\boldsymbol{X}_1^{(1)} = \begin{pmatrix} \boldsymbol{G}_1 \\ \boldsymbol{0} \\ \boldsymbol{0} \\ \boldsymbol{0} \end{pmatrix}, \quad \boldsymbol{X}_1^{(2)} = \begin{pmatrix} \boldsymbol{0} \\ \boldsymbol{G}_2 \\ \boldsymbol{0} \\ \boldsymbol{0} \end{pmatrix}, \quad \boldsymbol{X}_1^{(3)} = \begin{pmatrix} \boldsymbol{0} \\ \boldsymbol{0} \\ \boldsymbol{G}_3 \\ \boldsymbol{0} \end{pmatrix}, \quad \boldsymbol{X}_1^{(4)} = \begin{pmatrix} \boldsymbol{0} \\ \boldsymbol{0} \\ \boldsymbol{0} \\ \boldsymbol{G}_4 \end{pmatrix},$$

where each $\boldsymbol{G}_i$ is a matrix of size $n/4 \times (k+p)$. Here, one can obtain sketches of the blocks $\boldsymbol{H}_{21}^\top$ and $\boldsymbol{H}_{22}^\top$ in Eq. (17) using $\boldsymbol{X}_2^{(1)}$. With similar notation, six test vectors are needed to sample all the admissible blocks at the second level:

$$\boldsymbol{X}_2^{(1)} = \begin{pmatrix} \boldsymbol{G}_1 \\ \boldsymbol{0} \\ \boldsymbol{0} \\ \boldsymbol{0} \\ \boldsymbol{0} \\ \boldsymbol{0} \\ \boldsymbol{G}_{11} \\ \boldsymbol{0} \end{pmatrix}, \quad \boldsymbol{X}_2^{(2)} = \begin{pmatrix} \boldsymbol{0} \\ \boldsymbol{G}_6 \\ \boldsymbol{0} \\ \boldsymbol{0} \\ \boldsymbol{0} \\ \boldsymbol{0} \\ \boldsymbol{0} \\ \boldsymbol{G}_{12} \end{pmatrix}, \quad \boldsymbol{X}_2^{(3)} = \begin{pmatrix} \boldsymbol{0} \\ \boldsymbol{0} \\ \boldsymbol{G}_7 \\ \boldsymbol{0} \\ \boldsymbol{0} \\ \boldsymbol{0} \\ \boldsymbol{0} \\ \boldsymbol{0} \end{pmatrix}, \quad \boldsymbol{X}_2^{(4)} = \begin{pmatrix} \boldsymbol{0} \\ \boldsymbol{0} \\ \boldsymbol{0} \\ \boldsymbol{G}_8 \\ \boldsymbol{0} \\ \boldsymbol{0} \\ \boldsymbol{0} \\ \boldsymbol{0} \end{pmatrix}, \quad \boldsymbol{X}_2^{(4)} = \begin{pmatrix} \boldsymbol{0} \\ \boldsymbol{0} \\ \boldsymbol{0} \\ \boldsymbol{0} \\ \boldsymbol{G}_9 \\ \boldsymbol{0} \\ \boldsymbol{0} \\ \boldsymbol{0} \end{pmatrix}, \quad \boldsymbol{X}_2^{(4)} = \begin{pmatrix} \boldsymbol{0} \\ \boldsymbol{0} \\ \boldsymbol{0} \\ \boldsymbol{0} \\ \boldsymbol{0} \\ \boldsymbol{G}_{10} \\ \boldsymbol{0} \\ \boldsymbol{0} \end{pmatrix}.$$

From this level and until the stopping point, only six test vectors are needed at each level. The following corollary generalizes the error analysis for the peeling algorithm performed on a weakly admissible hierarchical matrix to a strongly admissible partition.

**Corollary 8** (Accumulation of errors in strongly admissible partitions). *Suppose the assumptions and notation in Proposition 7 hold for a hierarchical matrix $\boldsymbol{H}$ with a strongly admissible partition in dimension $d \geq 1$. Then, the sketch and projection errors during the peeling algorithm at level $1 \leq L \leq N$ satisfy the following bounds:*

$$\|\boldsymbol{Z}_L - \tilde{\boldsymbol{Z}}_L\|_{\mathrm{F}} \leq (6^d - 3^d) C \sqrt{k} \sum_{\ell=1}^{L-1} 2^{\frac{L-1-\ell}{2}} \tilde{\epsilon}_\ell, \quad \text{and} \quad \|\boldsymbol{Z}_L^P - \tilde{\boldsymbol{Z}}_L^P\|_{\mathrm{F}} \leq (6^d - 3^d) \sqrt{k} \sum_{\ell=1}^{L-1} 2^{\frac{L-1-\ell}{2}} \tilde{\epsilon}_\ell.$$

*Proof.* Let $1 \leq L \leq N$ be a hierarchical level. With the notation of Proposition 7, an input matrix $\boldsymbol{X}_L$ queries the action of a level-$L$ sub-block $\boldsymbol{H}(I_\alpha, I_\beta)$, where $\alpha$ and $\beta$ are level-$L$ nodes in the hierarchical tree and $I_\alpha, I_\beta \subset [0,1]^d$ are their corresponding index sets (see (26) for a description of index sets and hierarchical trees). Unfortunately, it also captures the actions of approximation errors in intersecting sub-blocks of $\boldsymbol{H} - \boldsymbol{H}^{(L-1)}$, due to using peeling at previous levels. As an example, if one tries to sketch the admissible block $\boldsymbol{H}_{421}^\top$ in Eq. (17) with a test vector, then the matrix-vector product might contain a contribution from $\boldsymbol{H}_{22}^\top - \tilde{\boldsymbol{H}}_{22}^\top$ and $\boldsymbol{H}_{24}^\top - \tilde{\boldsymbol{H}}_{24}^\top$. We derive an upper bound on the number of level-$\ell$ intersecting blocks for $1 \leq \ell \leq L-1$. Each of these intersecting blocks is the restriction of $\boldsymbol{H} - \boldsymbol{H}^{(L-1)}$ to an admissible block of the form $I_{\alpha^{(\ell)}} \times I_\beta$ from a previous level $\ell$, where $\alpha^{(\ell)}$ is the level-$\ell$ subdomain that was eventually subdivided to get $\alpha$. That is, $\alpha^{(L-1)}$ is the parent of the $\alpha$, $\alpha^{(L-2)}$ is the parent of $\alpha$'s parent, and so on. We then count the number of level-$\ell$ nodes $\beta$ for which $I_{\alpha^{(\ell)}} \times I_\beta$ is an admissible domain. First, $I_{\alpha^{(\ell)}} \times I_\beta$ is an admissible domain if and only if the nodes $\alpha^{(\ell)}$ and $\beta$ belong to each other's interaction lists (23). Moreover, the size of the interaction list of any node is bounded above by $6^d - 3^d$ (23), so this is also an

upper bound on the number of intersecting blocks from level $\ell$. Finally, each previously learned block from level $\ell$ contributes $\tilde{\epsilon}_\ell$ to the sum of accumulated errors for $1 \leq \ell \leq L-1$ (see Proposition 7), and we have just shown that there are at most $6^d - 3^d$ of such intersecting blocks at each level $\ell$. Combining this with Proposition 7, we derive the following bound on the perturbation error in the sketch step of the randomized SVD:

$$\|\boldsymbol{Z}_L - \tilde{\boldsymbol{Z}}_L\|_F \leq \sum_{\ell=1}^{L-1}(6^d - 3^d)\tilde{\epsilon}_\ell \sqrt{\sum_{j=1}^{2^{L-1-\ell}} \|\boldsymbol{x}_{2^{L-1-\ell}+j}\|_F^2} \leq (6^d - 3^d)C\sqrt{k}\sum_{\ell=1}^{L-1} 2^{\frac{L-1-\ell}{2}}\tilde{\epsilon}_\ell.$$

Similarly, we obtain a bound on the projection perturbation error:

$$\|\boldsymbol{Z}_L^P - \tilde{\boldsymbol{Z}}_L^P\|_F \leq \sum_{\ell=1}^{L-1}(6^d - 3^d)\tilde{\epsilon}_\ell \sqrt{\sum_{j=1}^{2^{L-1-\ell}} \|\boldsymbol{y}_{2^{L-1-\ell}+j}\|_F^2} \leq (6^d - 3^d)\sqrt{k}\sum_{\ell=1}^{L-1} 2^{\frac{L-1-\ell}{2}}\tilde{\epsilon}_\ell.$$

□

## 3. Stable recovery of Green's functions

This section generalizes the discussion of the randomized SVD and its perturbation analysis described in Section 2B to the continuous analogue of Hilbert–Schmidt operators, rather than matrices. We then use these results to derive the error analysis for the generalization of the peeling algorithm in infinite dimensions, showing that one can stably recover Green's functions on admissible domains. The key idea is to avoid the exponential accumulation of errors with each hierarchical level in the peeling procedure by taking advantage of the fast decay of a Green's function's singular values using a different target rank in the randomized SVD at each level. We then derive a probabilistic error bound for using infinite-dimensional peeling to approximate Green's functions on admissible domains from input-output pairs using the randomized SVD. Finally, we present a global error bound for the approximant in the operator norm that achieves a relative error of $\epsilon > 0$ on the entire domain.

**A. The randomized singular value decomposition for Hilbert–Schmidt operators.** The randomized SVD introduced in Section 2A has been recently generalized to compute low-rank approximants to Hilbert–Schmidt operators (13, 44). In this section, we consider the particular setting of using the randomized SVD to approximate a Green's function $G$ over an admissible domain $X \times Y \subset \Omega \times \Omega$. Analogous to sampling random input vectors, we sample random input functions defined on $Y$ from a Gaussian process $\mathcal{GP}(0, K_Y)$, where $K_Y : Y \times Y \to \mathbb{R}$ is the covariance kernel. We construct the secondary kernel $K_Y$ by transforming a global kernel $K : \Omega \times \Omega \to \mathbb{R}$, defined on the entire domain as follows. First, we assume that $K$ is a continuous symmetric positive-definite kernel with bounded trace, *i.e.*,

$$\text{Tr}(K) = \int_\Omega K(x,x)\,dx = \sum_{i=1}^\infty \lambda_i < \infty,$$

where $\lambda_1 \geq \lambda_2 \geq \cdots > 0$ are the eigenvalues of $K$. Using Mercer's theorem (70), there exists an orthonormal basis $\{e_i\}_i$ of $L^2(\Omega)$ of eigenfunctions of $K$, such that

$$K(x,y) = \sum_{i=1}^\infty \lambda_i e_i(x) e_i(y), \quad x, y \in \Omega,$$

where the convergence is absolute and uniform, and the eigenfunctions are continuous. We can then rescale, shift, restrict the $e_i$, and orthonormalize the family using Gram–Schmidt algorithm to generate an orthonormal family $\{e_{Y,i}\}_i$ in $L^2(Y)$. We then define the kernel $K_Y$ following its Mercer decomposition as

$$K_Y(x,y) = \sum_{i=1}^\infty \lambda_i e_{Y,i}(x) e_{Y,i}(y), \quad x, y \in Y.  \qquad [18]$$

This procedure guarantees that the secondary kernels have the same eigenvalues of $K$, and, in particular, that the trace is unchanged, which is crucial in the next sections to bound the norm of the random functions sampled from the associated GP.

As described in Section 1A, $G$ has low numerical rank on $X \times Y$, and the singular values associated with the corresponding Hilbert–Schmidt integral operator (see Eq. (2)) decay exponentially fast. Let $k \geq 1$ be a target rank and $k_\epsilon$ an oversampling parameter (see Section 1A). We sample random functions $f_1, \ldots, f_{k+k_\epsilon}$ from the Gaussian process $\mathcal{GP}(0, K_Y)$, and define the random quasimatrix[†] $\Omega_{X \times Y} = \begin{bmatrix} f_1 & \cdots & f_{k+k_\epsilon} \end{bmatrix}$. Analogous to the discrete case (43), we consider the $k \times (k + k_\epsilon)$ matrix $\Omega_{X \times Y, 1} = (\langle v_{i, X \times Y}, f_j \rangle)_{i,j}$, whose columns are i.i.d. and follow a multivariate Gaussian distribution with covariance matrix $\boldsymbol{C}_{X \times Y}$ (13, Lem. 1), defined as

$$[\boldsymbol{C}_{X \times Y}]_{ij} = \int_{Y \times Y} v_{i, X \times Y}(x) K_Y(x,y) v_{j, X \times Y}(y)\,dx\,dy, \quad 1 \leq i, j \leq k.$$

---

[†] See (71) for an introduction to the quasimatrix notation.

Here, $v_{i,X\times Y}$ is the $i$th singular vector of $G$ restricted to $X\times Y$. Following (13, Sec. 4.1.2), we sketch the HS operator at the random functions $f_1,\ldots,f_{k+k_\epsilon}$ and apply the randomized SVD to obtain an approximant $\tilde{G}$ to $G$ on the domain $X\times Y$. Then, with a target rank $k$ and an oversampling parameter $k_\epsilon$, the following error bound holds:

$$\|G-\tilde{G}_{X\times Y}\|^2_{L^2(X\times Y)} \leq \left(1+t^2s^2\frac{6k_\epsilon}{\gamma_{k,X\times Y}}\sum_{j=1}^{\infty}\frac{\lambda_j}{\lambda_1}\right)\epsilon^2\|G\|^2_{L^2(X\times\hat{Y})}$$

with probability $\geq 1-t^{-k_\epsilon}-e^{-s^2(k+k_\epsilon)}$ for any $s,t\geq 1$. The covariance kernel quality, $\gamma_{k,X\times Y}$, is defined as $\gamma_{k,X\times Y}=k/(\lambda_1\operatorname{Tr}(\boldsymbol{C}_{X\times Y}^{-1}))$. We can then choose $t=\mathcal{O}(1)$ and $s=\mathcal{O}(1)$ such that

$$\|G-\tilde{G}_{X\times Y}\|_{L^2(X\times Y)} = \mathcal{O}(k_\epsilon^{1/2}\gamma_{k,X\times Y}^{-1/2}\epsilon)\|G\|_{L^2(X\times\hat{Y})} \qquad [19]$$

holds with probability $\geq 1-e^{-k_\epsilon}$. We emphasize that the perturbation error analysis for the randomized SVD stated in Theorem 5 generalizes to infinite dimensions so that Eq. (19) holds whenever the magnitude of the perturbations is smaller than the target tolerance.

**B. Peeling for Green's functions.** We now generalize the peeling algorithm to recover Green's functions over hierarchically partitioned domains using the randomized SVD. In this section, we follow the model of HODLR matrix recovery algorithms (20, 69) to extend the peeling algorithm to the continuous case of Green's function recovery in dimensions $d\in\{1,2,3\}$. We recover $G$ over levels of the domain recursively, starting with the largest hierarchical level. First, the Green's function's behavior over all of the admissible domains at this level is recovered and stored in the function $G^{(1)}$. To get to the next level, we subdivide by halving the domain in all directions. Now, we seek to recover the function $G-G^{(1)}$ on any new admissible domains to obtain $G^{(2)}$. We dyadically partition any non-admissible domains and repeat the process to recover $G_2=G-G^{(1)}-G^{(2)}$ over the next level's admissible domains. We repeat until we reach a pre-determined stopping point close to the diagonal.

Let $1\leq L\leq N_\epsilon$ be a hierarchical level. We now describe the procedure for recovering $G_L=G-\sum_{\ell=1}^L G^{(\ell)}$ on each admissible subdomain of this level. The coloring algorithm easily applies to generating input functions over $\Omega$, as it was originally written in this generality. We first apply the graph coloring algorithm, described in Section 2D, to determine the chromatic number $\chi(G_L)$ associated with the different constraint sets for sketching $G_L$. As established in (23), we know that for the strongly admissible partition in dimension $d$, we have the following bound: $\chi(G_L)\leq 6^d$. Each color number $1\leq j\leq \chi(G_L)$ of the constraint graph gathers $n_j\geq 1$ constraints, which correspond to a set of admissible subdomains $\{X_i^j\times Y_i^j\}_{1\leq i\leq n_j}\subset\Omega\times\Omega$. We then construct test functions $f_j\in L^2(\Omega)$ such that

$$\operatorname{Supp}(f_j)\subset\cup_{i=1}^{n_j}Y_i^j,\quad\text{where}\quad f_j|_{Y_i^j}\sim\mathcal{GP}(0,K_{Y_i^j}),\quad 1\leq i\leq n_j,$$

and $K_{Y_i^j}$ is defined in Eq. (18). Given a target rank $k$ and an oversampling parameter $k_\epsilon$ for learning the Green's function with the randomized SVD over the admissible domains, we generate the input quasimatrix $F_L^{(j)}$, where $1\leq j\leq\chi(G_L)$, as follows:

$$F_L^{(j)} = \begin{bmatrix} f_{j,1} & \cdots & f_{j,k+k_\epsilon} \end{bmatrix},\quad\text{where}\quad f_{j,l}\sim f_j\quad\text{for}\quad 1\leq l\leq k+k_\epsilon.$$

Analogous to performing a matrix-vector product, sketching the Hilbert–Schmidt integral operator associated with $G_L$ with the quasimatrix $F_L^{(j)}$ yields $U_L^{(j)} = \begin{bmatrix} u_{j,1} & \cdots & u_{j,k+k_\epsilon} \end{bmatrix}$, where

$$u_{j,l}(x) = \int_\Omega G_L(x,y)f_{j,l}(y)\,\mathrm{d}y = \int_\Omega G(x,y)f_{j,l}(y)\,\mathrm{d}y - \sum_{\ell=1}^L\int_\Omega G^{(\ell)}(x,y)f_{j,l}(y)\,\mathrm{d}y,\quad 1\leq l\leq k+k_\epsilon.$$

We now restrict the output functions $u_{j,l}\in L^2(\Omega)$ to isolate the action of $G$ on the admissible subdomains $\{X_i^j\times Y_i^j\}_{1\leq i\leq n_j}$ associated with the $j$th color. Let $\mathcal{R}_{X_i^j}:L^2(\Omega)\to L^2(X_i^j)$ be the restriction operator to the subdomain $X_i^j\subset\Omega$. Then,

$$[\mathcal{R}_{X_i^j}u_{j,l}](x) = \mathcal{R}_{X_i^j}\int_\Omega G_L(x,y)f_{j,l}(y)\,\mathrm{d}y = \sum_{i'=1}^{n_j}\int_{Y_{i'}^j}\mathcal{R}_{X_i^j}G_L(x,y)f_{j,l}(y)\,\mathrm{d}y = \int_{Y_i^j}\mathcal{R}_{X_i^j}G(x,y)f_{j,l}|_{Y_i^j}(y)\,\mathrm{d}y,$$

where the final equality holds because the coloring algorithm ensures that there is only one $Y_i^j$ such that $X_i^j\times Y_i^j$ is an admissible domain. Any other domain of the form $X_i^j\times Y_{i'}^j$ for $i'\neq i$ belongs to a larger subdomain of $\Omega\times\Omega$ over which $G$ has already been recovered at a previous level. Thus, assuming exact recovery of peeling at previous hierarchical levels, $G_L$ is zero over such domains. Having isolated the action of $G_L$ on one admissible domain, we can now construct its approximation using the randomized SVD (13). Following this method for each color $1\leq j\leq \chi(G_L)$ of the graph coloring algorithm, we recover an approximation for $G$ over the admissible domains satisfying the corresponding sampling constraints. This results in a total of $2(k+k_\epsilon)\chi(G_L)\leq 4\times 6^d k_\epsilon$ input-output function pairs at each hierarchical level $1\leq L\leq N_\epsilon$. We emphasize that one application of the randomized SVD requires $2(k+k_\epsilon)$ pairs to perform the sketching and projection steps. Finally, if $X_j\times Y_j$ is an admissible domain at level $L$, with $1\leq j\leq N^{(L)}$, we denote by $\Omega_L^{(j)}\subset Y_j$ the associated random quasimatrix defined as

$$\Omega_L^{(j)} = \begin{bmatrix} f_{j,1}|_{Y_j} & \cdots & f_{j,k+k_\epsilon}|_{Y_j} \end{bmatrix},\quad f_{j,l}|_{Y_j}\sim\mathcal{GP}(0,K_{Y_j}). \qquad [20]$$

In practice, the number of random quasimatrices at level $L$ is much smaller than $N^{(L)}$, the number of admissible domains at that level.

**C. Error analysis of the infinite-dimensional peeling algorithm.** For a given target accuracy $\epsilon > 0$, we aim to reconstruct a Green's function $G$ corresponding to a self-adjoint Hilbert–Schmidt operator $H$, given in hierarchical format with $N_\epsilon \geq 1$ hierarchical levels in the domain and chromatic number bounded by $6^d$, to within relative error $\epsilon$. The reconstruction algorithm combines the randomized SVD and the peeling algorithm described in Sections 2C, 2D and 3B and only relies on sketches of $H$ with random test functions. In particular, if $\boldsymbol{H}$ is a matrix, we compute matrix-vector products using random input vectors sampled from a multivariate Gaussian distribution with zero mean and covariance matrix $K$. On the other hand, if $H : L^2(\Omega) \to L^2(\Omega)$ is a Hilbert–Schmidt operator in infinite dimensions, one can compute low-rank approximants using the HS randomized SVD described in Section 3A by evaluating the operator at random functions sampled from a Gaussian process with zero mean and the appropriate covariance kernel $K_Y$ described in Eq. (18).

More specifically, if $G_i$ represents the restriction of $G$ to some admissible domain $I_{\alpha_i} \times I_{\beta_i}$, where $I_{\alpha_i}, I_{\beta_i} \subset [0,1]^d$, then we define $H_i$ as the HS operator corresponding to the restriction $G_i$. We employ this notation for the rest of the section. Because $G$ is a Green's function, any such $H_i$ has exponentially fast decaying singular values. That is, for $0 < \epsilon < 1$, $H_i$ has numerical rank $k \leq k_\epsilon = M \log(1/\epsilon)^{d+1}$, where $d \geq 1$ is an integer and $M > 0$ is a constant independent of $\epsilon$. By the Eckart–Young theorem, the tail of the singular values of $H_i$ satisfies the following inequality:

$$\left( \sum_{j > k} \sigma_j(H_i)^2 \right)^{1/2} \leq \epsilon \|H\|_{\text{HS}}. \tag{21}$$

**Assumption 1** (Summary of assumptions). *We summarize the assumptions on the self-adjoint Hilbert-Schmidt operator $H$ as follows for ease of reference:*

1. *$H$ is a solution operator under the same assumptions as $\mathscr{F}$ in Eq. (2), with corresponding Green's function kernel $G$.*

2. *The domain of $G$ is hierarchically partitioned with $N_\epsilon \sim \frac{1}{2} \log_2(1/\epsilon)$ levels and chromatic number bounded by $6^d$.*

3. *For $\epsilon > 0$ sufficiently small, $H_i$ has numerical rank $k \leq k_\epsilon = M \log(1/\epsilon)^{d+1}$, where $M > 0$ is a constant.*

The following proposition gives a sufficient condition on the magnitude of the sketch perturbations to apply the perturbation estimate for the randomized SVD derived in Section 2B.

**Proposition 9** (Perturbed randomized SVD error). *Let the target tolerance $0 < \epsilon < 1$ be sufficiently small and $H_i$ be the restriction of $H$ to an admissible domain $X \times Y$. Suppose that sketching $H_i$ using a quasimatrix $\Omega_i$ with $k + k_\epsilon$ columns sampled i.i.d. from $\mathcal{GP}(0, K_Y)$ returns a perturbed sample $\tilde{Z} = H_i \Omega_i + E$, where the norm of the perturbation quasimatrix $E$ satisfies $\|E\|_{\text{HS}} \leq \gamma_k^{1/2} \sqrt{k + k_\epsilon} \epsilon^{1+\delta} \|H\|_{\text{HS}}$, for some $\delta > 0$. Then, under the condition that $\|\Omega_{i,1}\|_{\text{HS}} \geq \lambda_1 \gamma_k^{1/2} \epsilon^{\delta/2}$, the randomized SVD constructs an approximation $\tilde{H}_i$ of $H_i$ using $k + k_\epsilon$ input-output pairs satisfies*

$$\|H_i - \tilde{H}_i\|_{\text{HS}} = \mathcal{O}(\epsilon \sqrt{k_\epsilon/\gamma_k}) \|H\|_{\text{HS}},$$

*with probability $\geq 1 - e^{-k_\epsilon}$, where $k \leq k_\epsilon = M \log(1/\epsilon)^{d+1}$. Here, $\Omega_{i,1} = V_1^* \Omega_i \in \mathbb{R}^{k \times (k+k_\epsilon)}$, where $V_1$ is the quasimatrix containing the first $k$ right singular functions of $H_i$.*

*Proof.* The proof relies on the deterministic error bound in Proposition 4 for the randomized SVD with perturbed inputs. Here, one must control the decay rate of the $k$th singular value of $H_i$, denoted by $\sigma_k$, by providing a lower bound to ensure that the matrix $\boldsymbol{\Sigma}_1 \boldsymbol{E}_1$ in Eq. (11) has small norm so that one can perform a Taylor expansion of the pseudo-inverse. In the following, we argue that if the tail of the singular values of $H_i$ decays exponentially fast, then $\sigma_k$ must contribute to a fraction of the norm of $H_i$. For a given $0 < \epsilon < 1$, by combining (17, Thm. 2.8) and the Eckart–Young–Mirsky theorem for Hilbert–Schmidt operators, there exists $k \leq k_\epsilon = M \log(1/\epsilon)^{d+1}$ such that $H_i$ has numerical rank $k$. Let $\Sigma_2$ be the diagonal quasimatrix containing the singular values $\sigma_{k+1} \geq \sigma_{k+2} \geq \cdots \geq 0$ of $H_i$. We let $k \leq k_\epsilon$ be the unique integer satisfying the following two inequalities:

$$\|\Sigma_2\|_{\text{HS}}^2 \leq \epsilon^2 \|H\|_{\text{HS}}^2 \quad \text{and} \quad \sigma_k^2 + \|\Sigma_2\|_{\text{HS}}^2 > \epsilon^2 \|H\|_{\text{HS}}^2. \tag{22}$$

We aim to apply the perturbation analysis argument of Section 2B for the randomized SVD. We therefore must be able to estimate the term $\|\Sigma_1^{-1} E_1\|_{\text{HS}}$, where $\Sigma_1$ is the diagonal matrix containing the $k$ largest singular values of $H_i$. We apply the randomized SVD with a target rank of $k$ and an oversampling parameter of $k_\epsilon$, resulting in $k + k_\epsilon$ sketches of $H_i$ such that $E_1 = U_1^* E$ satisfies $\|E_1\|_{\text{HS}} \leq \|E\|_{\text{HS}} \leq \gamma_k^{1/2} \sqrt{k + k_\epsilon} \epsilon^{1+\delta} \|H\|_{\text{HS}} \leq \sqrt{2M} \gamma_k^{1/2} \log(1/\epsilon)^{\frac{d+1}{2}} \epsilon^{1+\delta} \|H\|_{\text{HS}}$. Then, we derive a lower bound on $\sigma_k$ to show that $\|\Sigma_1^{-1} E_1\|_{\text{HS}}$ converges to zero as $\epsilon$ goes to zero. We consider the following two cases:

1. If $\sigma_k \geq \epsilon \|H\|_{\text{HS}}/\sqrt{2}$, then $\|\Sigma_1^{-1}\|_{\text{HS}} \leq \sqrt{2k}/\epsilon \|H\|_{\text{HS}}$, and we have $\|\Sigma_1^{-1} E_1\|_{\text{HS}} \leq 2M \gamma_k^{1/2} \log(1/\epsilon)^{d+1} \epsilon^\delta$.

2. Otherwise, if $\sigma_k < \epsilon \|H\|_{\text{HS}}/\sqrt{2}$, then Eq. (22) implies that

$$\sum_{j > k} \sigma_j^2 > \frac{\epsilon^2}{2} \|H\|_{\text{HS}}^2. \tag{23}$$

Nicolas Boullé, Diana Halikias and Alex Townsend     **11 of 22**

We apply (17, Thm. 2.8) and the Eckart–Young–Mirsky theorem again with the accuracy $\epsilon/2$ to obtain a $k' \leq k_{\epsilon/2} = M\log(2/\epsilon)^{d+1} \leq 2^{d+1}k_\epsilon$ such that

$$\sum_{j>k'} \sigma_j^2 \leq \frac{\epsilon^2}{4}\|H\|_{\text{HS}}^2. \qquad [24]$$

We then combine Eqs. (23) and (24) to obtain the following lower bound on $\sigma_k$:

$$k'\sigma_k^2 \geq \sum_{j=k+1}^{k'} \sigma_j^2 = \sum_{j>k} \sigma_j^2 - \sum_{j>k'} \sigma_j^2 \geq \frac{\epsilon^2}{4}\|H\|_{\text{HS}}^2.$$

In the end, we find that $\|\Sigma_1^{-1}E_1\|_{\text{HS}} \leq 2^{d/2+2}M^{3/2}\gamma_k^{1/2}\log(1/\epsilon)^{3(d+1)/2}\epsilon^\delta$.

In both cases, we showed that $\|\Sigma_1^{-1}E_1\|_{\text{HS}} = o(\|\Omega_{i,1}\|_{\text{HS}})$. Then, we can perform a first order expansion for $\epsilon$ sufficiently small to obtain the randomized SVD bound (see Section 2B and Eq. (19)):

$$\|H_i - \tilde{H}_i\|_{\text{HS}} = \mathcal{O}(k_\epsilon^{1/2}\gamma_k^{-1/2}\epsilon)\|H\|_{\text{HS}},$$

which holds with probability $\geq 1 - e^{-k_\epsilon}$. We note that the error due to the projection step is negligible since we assumed this error is smaller than the target tolerance. □

Proposition 9 guarantees that one can apply the randomized SVD perturbation analysis whenever the magnitude of the perturbation is smaller than the target randomized SVD accuracy. We will now analyze the effect of the accumulation of errors in the peeling algorithm on the randomized SVD accuracy at the higher hierarchical levels. Following Proposition 7, the peeling algorithm introduces an accumulation of errors because admissible blocks from previous hierarchical levels perturb the sketches at the current level. To counterbalance this effect, we employ the randomized SVD at the first level with a higher target accuracy $\epsilon_1 = \epsilon^r$, where $r > 1$ is an exponent to be determined, and progressively decrease the target accuracy $\epsilon_L$ at hierarchical levels $1 \leq L \leq N_\epsilon$, such that $\epsilon_1 < \cdots < \epsilon_{N_\epsilon}$, to reach a relative error between the approximant and the Green's function of at most $\epsilon/\log(1/\epsilon)$ on each admissible subdomain $X \times Y \subset \Omega \times \Omega$ of the partition. Let $k_L \leq k_{\epsilon_L}$ be the target randomized SVD rank at level $1 \leq L \leq N_\epsilon$. We introduce the following covariance quality measure for the peeling algorithm:

$$\Gamma_\epsilon = \min_{1 \leq L \leq N_\epsilon} \left\{ \min_{X_L \times Y_L \text{ is admissible}} \gamma_{k_L, X_L \times Y_L} \right\}. \qquad [25]$$

We consider the following probability events, which provide bounds on the $j$th random quasimatrix at level $L$, $\Omega_L^{(j)}$, sampled from $\mathcal{GP}(0, K_{Y_j})$ (see Eq. (20)):

$$A_\Omega := \bigcap_{L=1}^{N_\epsilon} \bigcap_{j=1}^{N^{(L)}} \left\{ \|\Omega_L^{(j)}\|_{\text{HS}}^2 \leq 8k_{\epsilon_1}\text{Tr}(K) \right\}, \qquad B_\Omega := \bigcap_{L=1}^{N_\epsilon} \bigcap_{j=1}^{N^{(L)}} \left\{ \|\Omega_{L,1}^{(j)}\|_{\text{HS}}^2 \geq \lambda_1\epsilon^\delta\Gamma_\epsilon \right\}, \qquad [26]$$

where $\delta = 1/(N_\epsilon - 1)$, and $\Omega_{L,1}^{(j)}$ is the matrix containing the inner products of the first right singular vectors of $G$ restricted to the admissible domain $X \times Y$ and the random functions in $\Omega_L^{(j)}$ (see Section 3A). We analyze the probability that $A_\Omega$ and $B_\Omega$ occur in Section 3D. Then, Proposition 10 estimates the required exponent $r$ and decay rate of the target accuracy.

**Proposition 10** (Adaptive rate of the target accuracy). *Let $0 < \epsilon < 1$ sufficiently small and select the target accuracy for the randomized SVD at level $1 \leq L \leq N_\epsilon$ as follows:*

$$\epsilon_L = \mathcal{O}(k_{\epsilon_1}^{3/2}\Gamma_\epsilon^{-1}\epsilon^{-1/(N_\epsilon-1)}\epsilon_{L-1}), \quad \epsilon_1 = \epsilon^r,$$

*where $r = (d+2)\log(\log(1/\epsilon)) + \log(1/\Gamma_\epsilon)$, and $k_{\epsilon_1} = M\log(1/\epsilon_1)^{d+1}$. The factor $\Gamma_\epsilon$ is defined as the minimum of the covariance quality $\gamma_{\epsilon_L}$ over all hierarchical levels. Then, conditioning on $A_\Omega$, $B_\Omega$, and assuming that the randomized SVD succeeds on each admissible block, the error between the Green's function and its approximant $\tilde{G}$ returned by Algorithm 1 using the randomized SVD and the peeling algorithm satisfies*

$$\|G - \tilde{G}\|_{L^2(X \times Y)} \leq \frac{\epsilon}{\log(1/\epsilon)}\|G\|_{L^2(\Omega \times \Omega)},$$

*on each admissible domain $X \times Y \subset \Omega \times \Omega$ of the hierarchical partition.*

*Proof.* Let $1 \leq L \leq N_\epsilon$ be a hierarchical level and denote by $\tilde{\epsilon}_\ell$ the absolute error between the Green's function and its approximant on each of the admissible domains at the previous levels $1 \leq \ell \leq L-1$. According to Corollary 8, the approximation error of the perturbed sketches (randomized SVD sketch and projection sketch) at level $L$ are bounded by $E_L$ and $E_L^P$ as

$$\|Z_L - \tilde{Z}_L\|_F \leq E_L, \quad E_L := \sqrt{2}(6^d - 3^d)C\sqrt{k_{\epsilon_1}}\sum_{\ell=1}^{L-1} 2^{\frac{L-1-\ell}{2}}\tilde{\epsilon}_\ell = \sqrt{2}E_{L-1} + \sqrt{2}(6^d - 3^d)C\sqrt{k_{\epsilon_1}}\tilde{\epsilon}_{L-1}, \qquad [27a]$$

$$\|Z_L^P - \tilde{Z}_L^P\|_F \leq E_L^P, \quad E_L^P := \sqrt{2}(6^d - 3^d)\sqrt{k_{\epsilon_1}}\sum_{\ell=1}^{L-1} 2^{\frac{L-1-\ell}{2}}\tilde{\epsilon}_\ell = \sqrt{2}E_{L-1}^P + \sqrt{2}(6^d - 3^d)\sqrt{k_{\epsilon_1}}\tilde{\epsilon}_{L-1}, \qquad [27b]$$

where $C = 2\sqrt{2}\operatorname{Tr}(K)^{1/2}k_{\epsilon_1}^{1/2}$ following Eq. (26). Note that we bounded the target rank $k$ at level $\ell$ by the initial largest target rank $k_{\epsilon_1}$ since we start the peeling algorithm with the smallest target accuracy $\epsilon^r = \epsilon_1 \leq \cdots \leq \epsilon_{N_\epsilon}$. Then, Eq. (27) yields

$$\max\{E_L, E_L^P\} \leq \sqrt{2}\max\{E_{L-1}, E_{L-1}^P\} + 4(6^d - 3^d)\operatorname{Tr}(K)^{1/2}k_{\epsilon_1}\tilde{\epsilon}_{L-1}. \qquad [28]$$

Now, following Proposition 9, one can apply the randomized SVD perturbation analysis whenever the sketch perturbations are asymptotically smaller than the target accuracy. That is, if $E_L$ and $E_L^P$ satisfy $\max\{E_L, E_L^P\} = \Gamma_\epsilon^{1/2}\epsilon^\delta\epsilon_L$ for some $\delta > 0$. The target randomized SVD accuracy, $\epsilon_L$, and resulting absolute approximation error, $\tilde{\epsilon}_L$, are related by the following equation (see Proposition 9):

$$\tilde{\epsilon}_L = \mathcal{O}(k_{\epsilon_L}^{1/2}\gamma_{\epsilon_L}^{-1/2}\epsilon_L) = \mathcal{O}(k_{\epsilon_1}^{1/2}\Gamma_\epsilon^{-1/2}\epsilon_L), \qquad [29]$$

where $k_{\epsilon_L} = M\log(1/\epsilon_L)^{d+1} \leq k_{\epsilon_1}$ and $\gamma_{\epsilon_L}^{-1/2} \leq \Gamma_\epsilon^{-1/2}$. Combining Eqs. (28) and (29) yields

$$\epsilon_L = \mathcal{O}\left(\frac{\sqrt{2}\epsilon^\delta + 4(6^d - 3^d)\operatorname{Tr}(K)^{1/2}k_{\epsilon_1}^{3/2}\Gamma_\epsilon^{-1}}{\epsilon^\delta}\epsilon_{L-1}\right) = \mathcal{O}(k_{\epsilon_1}^{3/2}\Gamma_\epsilon^{-1}\epsilon^{-\delta}\epsilon_{L-1}). \qquad [30]$$

Therefore, after iterating Eq. (30) over $1 \leq \ell \leq L$, we obtain the following estimate for the target randomized SVD accuracy at level $L$: $\epsilon_L = \mathcal{O}(k_{\epsilon_1}^{3(L-1)/2}\Gamma_\epsilon^{-(L-1)}\epsilon^{-(L-1)\delta}\epsilon_1)$. Moreover, the initial randomized SVD target rank is bounded by $k_{\epsilon_1} = M\log(1/\epsilon_1)^{d+1} = Mr^{d+1}\log(1/\epsilon)^{d+1}$ since $\epsilon_1 = \epsilon^r$ by definition. Finally, with a choice of $\delta = 1/(N_\epsilon - 1)$, at the final level $N_\epsilon \sim \log(1/\epsilon)/(2\log 2)$, the absolute accuracy satisfies

$$\tilde{\epsilon}_{N_\epsilon} = \mathcal{O}\left(\epsilon^{r-1}k_{\epsilon_1}^{3N_\epsilon/2}\Gamma_\epsilon^{-N_\epsilon}\right) = \mathcal{O}\left(\epsilon^{r-1-\frac{1}{2\log 2}[\frac{3}{2}\log(k_{\epsilon_1}) + \log(1/\Gamma_\epsilon)]}\right).$$

Then, we select $r$ such that $\tilde{\epsilon}_{N_\epsilon} = \mathcal{O}(\epsilon/\log(1/\epsilon))$, *i.e.*,

$$r - 1 - \frac{3}{4\log 2}\log(M) - \frac{3}{4\log 2}(d+1)\log(r) - \frac{3}{4\log 2}(d+1)\log(\log(1/\epsilon)) - \frac{1}{2\log 2}\log(1/\Gamma_\epsilon) \geq 1 + \frac{\log(\log(1/\epsilon))}{\log(1/\epsilon)}. \qquad [31]$$

We then choose $r = (d+2)\log(\log(1/\epsilon)) + \log(1/\Gamma_\epsilon)$ so that Eq. (31) holds for $\epsilon$ sufficiently small, which concludes the proof. $\square$

Proposition 10 estimates the randomized SVD target accuracy, $\epsilon_L$, at each hierarchical level needed to recover the Green's function to within $\epsilon/\log(1/\epsilon)$ relative error. One can then provide an upper bound on the number of input-output pairs needed by Algorithm 1 to approximate the Green's function on each admissible domain.

**Corollary 11** (Number of input-output pairs). *Assume that the randomized SVD is successful on each admissible block of the Green's function up to level $N_\epsilon$. Then, the total number of input-output pairs required to approximate each off-diagonal block of the Green's function to within relative error $\epsilon/\log(1/\epsilon)$ is bounded by $\mathcal{O}(\log(1/\epsilon)^{d+2}[\log(\log(1/\epsilon)) + \log(1/\Gamma_\epsilon)]^{d+1})$ as $\epsilon \to 0$.*

*Proof.* Following Proposition 10, we choose a target rank and an oversampling parameter for the randomized SVD bounded by $k_{\epsilon_L} \leq k_{\epsilon_1} = M\log(1/\epsilon_1)^{d+1}$ at each hierarchical level. Moreover, for a given level $1 \leq L \leq N_\epsilon$, the randomized SVD requires a number of input-output pairs that is bounded by $4k_{\epsilon_L}\chi(G_L) \leq 4 \times 6^d k_{\epsilon_1}$ to approximate the Green's function on all the level-$L$ admissible subdomains (see Section 3B). Then, the total number of input-output pairs after $N_\epsilon \sim \log(1/\epsilon)/(2\log 2)$ hierarchical levels is bounded by

$$N \leq 4 \times 6^d N_\epsilon k_{\epsilon_1} = \mathcal{O}(r^{d+1}\log(1/\epsilon)^{d+2}) = \mathcal{O}(\log(1/\epsilon)^{d+2}[\log(\log(1/\epsilon)) + \log(1/\Gamma_\epsilon)]^{d+1}).$$

$\square$

**D. Probabilistic error analysis.** The previous section provided an upper bound on the number of input-output pairs required by Algorithm 1 to construct an approximant to the Green's function, conditionally upon the success of the randomized SVD at each hierarchical level. In this section, we provide a lower bound on the probability of success of Algorithm 1. We begin with the following lemma (see (13, Lem. 4) for a proof), which gives a Chernoff-type bound (72) for the HS-norm of the random input functions. This controls the constant $C$ in Proposition 7 to bound the norm of the sketch error during peeling.

**Lemma 12** (Chernoff bound for Gaussian processes). *Let $\Omega_{X \times Y}$ be a random quasimatrix with $\ell \geq 1$ i.i.d. columns in $L^2(Y)$ for $X \times Y \subset \Omega \times \Omega$, with each column following a Gaussian process $\mathcal{GP}(0, K_Y)$, where $K_Y$ is defined in Eq. (18). For all $s \geq 1$,*

$$\mathbb{P}\left\{\|\Omega_{X \times Y}\|_{\mathrm{HS}}^2 \leq \ell s^2 \operatorname{Tr}(K)\right\} \geq 1 - \left[se^{-(s^2-1)/2}\right]^\ell,$$

*where $\operatorname{Tr}(K) = \operatorname{Tr}(K_Y)$ is the sum of the eigenvalues of the kernel $K$.*

We can then bound the norm of the random quasimatrices used in the peeling algorithm of Section 3C with Lemma 12 and estimate the probability of the event $A_\Omega$ defined in Eq. (26).

**Proposition 13** (Global upper bound of the forcing terms). *Let $1 \leq L \leq N_\epsilon$ be a hierarchical level and $1 \leq j \leq N^{(L)} \leq 6^d 2^{dL}$, where $N^{(L)}$ is the number of admissible domains at level $L$. Let $\Omega_L^{(j)}$ be a random quasimatrix used at level $L$ of the peeling algorithm with i.i.d. columns defined in Section 3B. Then,*

$$\mathbb{P}(A_\Omega) = \mathbb{P}\left(\bigcap_{L=1}^{N_\epsilon} \bigcap_{j=1}^{N^{(L)}} \left\{\|\Omega_L^{(j)}\|_{\mathrm{HS}}^2 \leq 8 k_{\epsilon_1} \operatorname{Tr}(K)\right\}\right) \geq 1 - \frac{1}{4} e^{-\log(1/\epsilon)^d}.$$

*Proof.* Let $k_L \leq k_{\epsilon_L}$ be the target rank at level $L$ such that the quasimatrix $\Omega_L^{(j)}$ has $k_L + k_{\epsilon_L}$ columns, where $k_{\epsilon_L} = M \log(1/\epsilon_L)^{d+1}$ is the oversampling parameter. Applying Lemma 12 for $s \geq 2$ yields

$$\mathbb{P}\left\{\|\Omega_L^{(j)}\|_{\mathrm{HS}}^2 \leq (k_L + k_{\epsilon_L}) s^2 \operatorname{Tr}(K)\right\} \geq 1 - \left[s e^{-(s^2-1)/2}\right]^{k_L + k_{\epsilon_L}} \geq 1 - \left[s e^{-(s^2-1)/2}\right]^{k_\epsilon},$$

as $k_\epsilon \leq k_{\epsilon_L}$ for $1 \leq L \leq N_\epsilon$ and $\operatorname{Tr}(K) = \operatorname{Tr}(K_Y)$. We denote by $A_\Omega^{(s)}$ the event

$$A_\Omega^{(s)} = \bigcap_{L=1}^{N_\epsilon} \bigcap_{j=1}^{N^{(L)}} \left\{\|\Omega_L^{(j)}\|_{\mathrm{HS}}^2 \leq (k_L + k_{\epsilon_L}) s^2 \operatorname{Tr}(K)\right\}.$$

We can compute a lower bound for $\mathbb{P}(A_\Omega^{(s)})$ using the independence of the random samples from the Gaussian process as

$$\mathbb{P}(A_\Omega^{(s)}) = \prod_{L=1}^{N_\epsilon} \prod_{j=1}^{N^{(L)}} \mathbb{P}\left\{\|\Omega_L^{(j)}\|_{\mathrm{HS}}^2 \leq (k_L + k_{\epsilon_L}) s^2 \operatorname{Tr}(K)\right\} \geq \left(1 - \left[s e^{-(s^2-1)/2}\right]^{k_\epsilon}\right)^{6^d 2^{d(N_\epsilon+1)}}$$
$$\geq \exp\left\{6^d 2^{d(N_\epsilon+1)} \log\left(1 - e^{-k_\epsilon[(s^2-1)/2 - \log(s)]}\right)\right\}, \quad [32]$$

where $k_\epsilon[(s^2-1)/2 - \log(s)] \to \infty$ as $\epsilon \to 0$ since $(s^2-1)/2 > \log(s)$ for $s \geq 2$. Let $s = 2$, $k_\epsilon = M \log(1/\epsilon)^{d+1}$, and define the constant $C_1 = M[3/2 - \log(2)] > 0$ such that

$$\log\left(1 - e^{-k_\epsilon[3/2 - \log(2)]}\right) = \log\left(1 - e^{-C_1 \log(1/\epsilon)^{d+1}}\right) \geq -2 e^{-C_1 \log(1/\epsilon)^{d+1}}, \quad [33]$$

where we used the inequality $\log(1-u) \geq -u/(1-u) \geq -2u$ for $|u| < 1/2$. Moreover, since $N_\epsilon \sim \log(1/\epsilon)/(2\log 2)$ (see Eq. (7)), there exists a constant $C_2 > 0$ such that $6^d 2^{d(N_\epsilon+1)} \leq C_2 \epsilon^{-2d}$ for sufficiently small $\epsilon$. Therefore, combining Eqs. (32) and (33) yields

$$\mathbb{P}(A_\Omega^{(2)}) \geq \exp\left\{-C_2 e^{-C_1 \log(1/\epsilon)^{d+1} + 2d \log(1/\epsilon)}\right\} \geq \exp\left\{-\frac{1}{4} e^{-\log(1/\epsilon)^d}\right\} \geq 1 - \frac{1}{4} e^{-\log(1/\epsilon)^d},$$

where we used the fact that $C_2 e^{-C_1 \log(1/\epsilon)^{d+1} + 2d \log(1/\epsilon)} \leq e^{-\log(1/\epsilon)^d}/4$ for sufficiently small $\epsilon$, and the inequality $e^{-u} \geq 1 - u$ for $|u| < 1$. We then obtain the following bound,

$$\mathbb{P}\left(\bigcap_{L=1}^{N_\epsilon} \bigcap_{j=1}^{N^{(L)}} \left\{\|\Omega_L^{(j)}\|_{\mathrm{HS}}^2 \leq 4(k_L + k_{\epsilon_L}) \operatorname{Tr}(K)\right\}\right) \geq 1 - \frac{1}{4} e^{-\log(1/\epsilon)^d}.$$

Finally, we note that bounding $\|\Omega_L^{(i)}\|_{\mathrm{HS}}^2$ by $4(k_L + k_{\epsilon_L}) \operatorname{Tr}(K)$ implies the inequality $\|\Omega_L^{(i)}\|_{\mathrm{HS}}^2 \leq 8 k_{\epsilon_1} \operatorname{Tr}(K)$, because $k_L \leq k_{\epsilon_L} \leq k_{\epsilon_1}$, which achieves the proof. □

The next lemma gives a lower bound on the Frobenius norm of a matrix with i.i.d. columns sampled from a multivariate Gaussian distribution.

**Lemma 14** (Chernoff lower bound). *Let $\boldsymbol{\Omega}_1 \in \mathbb{R}^{k \times \ell}$ be a random matrix with $\ell \geq k \geq 1$, where each column is sampled from a multivariate Gaussian distribution with mean zero and covariance matrix $\boldsymbol{C} \in \mathbb{R}^{k \times k}$. Then, for $0 < c < 1$ we have*

$$\mathbb{P}(\|\boldsymbol{\Omega}_1\|_{\mathrm{HS}}^2 \geq c \lambda_k(\boldsymbol{C}) \ell) \geq 1 - \left[e^{-\log(c) + c - 1}\right]^{-\ell/2},$$

*where $\lambda_k(\boldsymbol{C})$ is the smallest eigenvalue of $\boldsymbol{C}$.*

*Proof.* Let $Z = \|\Omega_1\|_{\text{HS}}^2 = \text{Tr}(A)$, where the scatter matrix $A = \Omega_1\Omega_1^*$ follows the Wishart distribution $W_k(C, \ell)$ with scale matrix $C \in \mathbb{R}^{k \times k}$. We denote the eigenvalues of $C$ by $\lambda_1(C) \geq \cdots \geq \lambda_k(C)$. Following (73, Sec. 3), the moment generating function of $Z$ is

$$\mathbb{E}[e^{tZ}] = \det(I - 2tC)^{-\ell/2}.$$

Let $c > 0$ and $t < 0$, using Markov's inequality we have

$$\mathbb{P}(Z < c\lambda_k(C)\ell) = \mathbb{P}(e^{tZ/\lambda_k(C)} > e^{tc\ell}) \leq \det\left(I - \frac{2t}{\lambda_k(C)}C\right)^{-\ell/2} e^{-tc\ell} = e^{-\ell\left[\frac{1}{2}\log\det\left(I - \frac{2t}{\lambda_k(C)}C\right) + tc\right]}$$

$$\leq e^{-\ell\left[\frac{k}{2}\log(1-2t) + tc\right]},$$

[34]

where the last inequality comes from the following relation:

$$\log\det\left(I - \frac{2t}{\lambda_k(C)}C\right) = \sum_{i=1}^k \log\left(1 - 2t\frac{\lambda_i(C)}{\lambda_k(C)}\right) \geq k\log(1 - 2t).$$

We then choose $t = 1 - k/c$ to minimize the right-hand side of Eq. (34), which yields

$$\mathbb{P}(Z < c\lambda_k(C)\ell) \leq \left[e^{\log(k/c) + (c/k - 1)}\right]^{-\ell k/2} \leq \left[e^{\log(c) + 1 - c}\right]^{\ell/2}.$$

$\square$

We can now estimate the probability of $B_\Omega$ in Eq. (26) using Lemma 14.

**Proposition 15** (Global lower bound of the forcing terms). *Let $1 \leq L \leq N_\epsilon$ be a hierarchical level and $1 \leq j \leq N^{(L)} \leq 6^d 2^{dL}$, where $N^{(L)}$ is the number of admissible domains at level $L$. Let $\Omega_L^{(j)}$ be a random quasimatrix used at level $L$ of the peeling algorithm with i.i.d. columns defined in Section 3B and $\Omega_{L,1}^{(j)}$ be the matrix containing the inner products with right singular vectors of $G$ restricted to the corresponding admissible domain (see Section 3A). Then,*

$$\mathbb{P}(B_\Omega) = \mathbb{P}\left(\bigcap_{L=1}^{N_\epsilon} \bigcap_{j=1}^{N^{(L)}} \left\{\|\Omega_{L,1}^{(j)}\|_{\text{HS}}^2 \geq \lambda_1 \epsilon^\delta \Gamma_\epsilon\right\}\right) \geq 1 - \frac{1}{4}e^{-\log(1/\epsilon)^d},$$

*where $\delta = 1/(N_\epsilon - 1)$ (see Proposition 10).*

*Proof.* Let $1 \leq L \leq N_\epsilon$, $1 \leq j \leq N^{(L)}$, and $\Omega_{L,1}^{(j)} \in \mathbb{R}^{k_L \times (k_L + k_{\epsilon_L})}$ be the random matrix associated with the quasimatrix $\Omega_L^{(j)}$. Let $0 < c < 1$, following Lemma 14, we have

$$\mathbb{P}\left(\|\Omega_{L,1}^{(j)}\|_{\text{HS}}^2 \geq c\lambda_1 \Gamma_\epsilon\right) \geq 1 - e^{k_\epsilon [\log(c) - c + 1]/2},$$

since $\lambda_{k_L}(C)(k_L + k_{\epsilon_L}) \geq \lambda_1 \Gamma_\epsilon$ and $k_L + k_{\epsilon_L} \geq k_\epsilon$. Therefore, choosing $c = \epsilon^\delta$ yields

$$\mathbb{P}(B_\Omega) = \mathbb{P}\left(\bigcap_{L=1}^{N_\epsilon} \bigcap_{j=1}^{N^{(L)}} \left\{\|\Omega_{L,1}^{(j)}\|_{\text{HS}}^2 \geq \lambda_1 \epsilon^\delta \Gamma_\epsilon\right\}\right) \geq \left(1 - e^{k_\epsilon[-\delta\log(1/\epsilon) - \epsilon^\delta + 1]/2}\right)^{6^d 2^{d(N_\epsilon+1)}}$$

$$\geq \exp\left\{6^d 2^{d(N_\epsilon+1)} \log\left(1 - e^{k_\epsilon[-\delta\log(1/\epsilon) - \epsilon^\delta + 1]/2}\right)\right\} \geq \exp\left\{-6^d 2^{d(N_\epsilon+1)+1} e^{k_\epsilon[-\delta\log(1/\epsilon) - \epsilon^\delta + 1]/2}\right\},$$

where we used the inequality $\log(1 - u) \geq -2u$ for $|u| < 1/2$. Let $C_2 > 0$ be a constant such that $6^d 2^{d(N_\epsilon+1)} \leq C_2 \epsilon^{-2d}$ for sufficiently small $\epsilon$. Then,

$$\mathbb{P}(B_\Omega) \geq \exp\left\{-2C_2 e^{2d\log(1/\epsilon) - M\log(1/\epsilon)^{d+1}(\delta\log(1/\epsilon) + \epsilon^\delta - 1)}\right\} \geq \exp\left\{-\frac{1}{4}e^{-\log(1/\epsilon)^d}\right\} \geq 1 - \frac{1}{4}e^{-\log(1/\epsilon)^d},$$

where we used the fact that $2C_2 e^{2d\log(1/\epsilon) - M\log(1/\epsilon)^{d+1}(\delta\log(1/\epsilon) + \epsilon^\delta - 1)} \leq e^{-\log(1/\epsilon)^d}/4$ for sufficiently small $\epsilon$, and the inequality $e^{-u} \geq 1 - u$ for $|u| < 1$. $\square$

We combine Proposition 13 and the probability bounds for the Hilbert–Schmidt randomized SVD (cf. Section 3A) to obtain a global probability bound for the algorithm that uses the peeling procedure with the randomized SVD.

**Proposition 16** (Probabilistic bound). *Let $0 < \epsilon < 1$ be sufficiently small. There is a randomized algorithm that constructs an approximation to the Green's function using $N = \mathcal{O}(\log(1/\epsilon)^{d+2}[\log(\log(1/\epsilon)) + \log(1/\Gamma_\epsilon)]^{d+1})$ input-output pairs $\{(f_j, u_j)\}_{j=1}^N$ such that*

$$\|G - \tilde{G}\|_{L^2(X \times Y)} \leq \frac{\epsilon}{\log(1/\epsilon)} \|G\|_{L^2(\Omega \times \Omega)},$$

[35]

*holds on all admissible domains $X \times Y$ in the hierarchical partition of $G$ with $N_\epsilon$ levels with probability $\geq 1 - e^{-\log(1/\epsilon)^d}$.*

*Proof.* We first note that Eq. (35) holds if the randomized SVD is successful at all the admissible blocks of the hierarchical partition of the Green's function. Therefore, if $\mathcal{E}_\epsilon$ denotes the following event:

$$\mathcal{E}_\epsilon = \{\|G - \tilde{G}\|_{L^2(X \times Y)} \leq \frac{\epsilon}{\log(1/\epsilon)} \|G\|_{L^2(\Omega \times \Omega)}, \text{ for all admissible domains } X \times Y\},$$

then we have $\mathbb{P}(\mathcal{E}_\epsilon) \geq \mathbb{P}(\cap_{L=1}^{N_\epsilon} A_L)$, where $A_L$ is the event that all the applications of the randomized SVD are successful for the admissible domains in the level $1 \leq L \leq N_\epsilon$ of the hierarchical partition. Since the probability of success of the randomized SVD depends on the norm of the random input vectors due to the peeling procedure, we condition on the event $A_\Omega \cap B_\Omega$, where again $A_\Omega$ and $B_\Omega$ are defined in Eq. (26), to obtain

$$\mathbb{P}(\mathcal{E}_\epsilon) \geq \mathbb{P}\left(\bigcap_{L=1}^{N_\epsilon} A_L \mid A_\Omega \cap B_\Omega\right) \mathbb{P}(A_\Omega \cap B_\Omega). \qquad [36]$$

By Propositions 13 and 15, we have the inequalities $\mathbb{P}(A_\Omega) \geq 1 - e^{-\log(1/\epsilon)^d}/4$ and $\mathbb{P}(B_\Omega) \geq 1 - e^{-\log(1/\epsilon)^d}/4$. We therefore focus on deriving a lower bound for $\mathbb{P}(\cap_{L=1}^{N_\epsilon} A_L \mid A_\Omega \cap B_\Omega)$.

Let $2 \leq L \leq N_\epsilon$ be a hierarchical level. As in Section 3C, due to the peeling procedure and accumulation of the perturbation errors, the probability of success of the randomized SVD bound given by Proposition 9 on the admissible blocks at level $L$ depends on the success of the randomized SVDs at all previous levels $1 \leq \ell \leq L - 1$. If $p_L = \mathbb{P}(\cap_{\ell=1}^{L} A_\ell \mid A_\Omega \cap B_\Omega)$ denotes the probability of success of the randomized SVD at each level up to $L$ given $A_\Omega \cap B_\Omega$, we can estimate $p_L$ using conditional probability as follows:

$$p_L = \mathbb{P}\left(\bigcap_{\ell=1}^{L} A_\ell \mid A_\Omega \cap B_\Omega\right) = \mathbb{P}\left(A_L \mid \bigcap_{\ell=1}^{L-1} A_\ell, A_\Omega \cap B_\Omega\right) \mathbb{P}\left(\bigcap_{\ell=1}^{L-1} A_\ell \mid A_\Omega \cap B_\Omega\right) = \mathbb{P}\left(A_L \mid \bigcap_{\ell=1}^{L-1} A_\ell, A_\Omega \cap B_\Omega\right) p_{L-1}, \quad [37]$$

where $p_1 = \mathbb{P}(A_1 \mid A_\Omega \cap B_\Omega)$. We can now use the HS randomized SVD probability bound (see Eq. (19)) to derive a lower bound for $p_L$. First, denote by $N^{(L)}$ the number of admissible domains at level $L$, and consider the event $B_L^i$ that the approximation of $G$ obtained by the randomized SVD on the $i$th admissible domain $X \times Y$ with target accuracy $\epsilon_L > 0$ satisfies Eq. (19), i.e.,

$$B_L^i = \left\{\|G - \tilde{G}\|_{L^2(X \times Y)} \leq \mathcal{O}\left(k_{\epsilon_L}^{1/2} \Gamma_\epsilon^{-1/2} \epsilon_L\right) \|G\|_{L^2(\Omega \times \Omega)}\right\}.$$

This is an application of the randomized SVD with a target rank of $k_L \leq k_{\epsilon_L}$ and an oversampling parameter of $k_{\epsilon_L} \geq k_\epsilon$, where $k_\epsilon = M \log(1/\epsilon)^{d+1}$. Therefore, using the HS randomized SVD probability bound (see Eq. (19)), we have

$$\mathbb{P}\left(B_L^i \mid \bigcap_{\ell=1}^{L-1} A_\ell, A_\Omega \cap B_\Omega\right) \geq 1 - e^{-k_\epsilon}. \qquad [38]$$

Moreover, combining Eqs. (37) and (38), and using Boole's inequality for the union of events yields

$$p_L = \mathbb{P}\left(\bigcap_{i=1}^{N^{(L)}} B_L^i \mid \bigcap_{\ell=1}^{L-1} A_\ell, A_\Omega \cap B_\Omega\right) p_{L-1} = \left(1 - \mathbb{P}\left(\bigcup_{i=1}^{N^{(L)}} \bar{B}_L^i \mid \bigcap_{\ell=1}^{L-1} A_\ell, A_\Omega \cap B_\Omega\right)\right) p_{L-1}$$
$$\geq \left(1 - \sum_{i=1}^{N^{(L)}} \mathbb{P}\left(\bar{B}_L^i \mid \bigcap_{\ell=1}^{L-1} A_\ell, A_\Omega \cap B_\Omega\right)\right) p_{L-1} \geq (1 - N^{(L)} e^{-k_\epsilon}) p_{L-1}, \qquad [39]$$

where $\bar{B}_L^i$ denotes the complementary event of $B_L^i$ such that $\mathbb{P}(\bar{B}_L^i \mid \cap_{\ell=1}^{L-1} A_\ell, A_\Omega \cap B_\Omega) \leq e^{-k_\epsilon}$, and $p_1 \geq 1 - N^{(1)} e^{-k_\epsilon}$ by Eq. (38). Iterating Eq. (39) over $1 \leq L \leq N_\epsilon$ gives the following probability bound:

$$p_{N_\epsilon} \geq \prod_{L=1}^{N_\epsilon} (1 - N^{(L)} e^{-k_\epsilon}), \qquad [40]$$

where the number of admissible domains at level $L$ is bounded by $N^{(L)} \leq 6^d 2^{dL}$. Therefore, we take the logarithm of Eq. (40) to determine a lower bound on $\log(p_{N_\epsilon})$,

$$\log(p_{N_\epsilon}) \geq \sum_{L=1}^{N_\epsilon} \log(1 - N^{(L)} e^{-k_\epsilon}) = -\sum_{L=1}^{N_\epsilon} \sum_{i=1}^{\infty} \frac{N^{(L)i} e^{-ik_\epsilon}}{i} \geq -\sum_{i=1}^{\infty} \frac{6^{di} e^{-ik_\epsilon}}{i} \sum_{L=1}^{N_\epsilon} 2^{idL} = -\sum_{i=1}^{\infty} \frac{6^{di} e^{-ik_\epsilon}}{i} \frac{2^{id(N_\epsilon+1)} - 1}{2^{id} - 1}$$
$$\geq -\sum_{i=1}^{\infty} \frac{6^{di} 2^{id(N_\epsilon+1)} e^{-ik_\epsilon}}{i} = \log(1 - 6^d 2^{d(N_\epsilon+1)} e^{-k_\epsilon}). \qquad [41]$$

Here, we used the Taylor series of $\log(1-u)$ for $|u| < 1$. Moreover, for sufficiently small $\epsilon$, we have $6^d 2^{d(N_\epsilon+1)} e^{-k_\epsilon} \leq C_1 \epsilon^{-d} e^{-M \log(1/\epsilon)^{d+1}}$ for some constant $C_1 > 0$. Finally, taking the exponential of Eq. (41) gives,

$$p_{N_\epsilon} \geq 1 - C_1 \epsilon^{-d} e^{-M \log(1/\epsilon)^{d+1}} = 1 - C_1 e^{-\log(1/\epsilon)[M \log(1/\epsilon)^d - d]} \geq 1 - \frac{1}{2} e^{-\log(1/\epsilon)^d},$$

for $\epsilon$ sufficiently small. We conclude with Eq. (36) as

$$\mathbb{P}(\mathcal{E}_\epsilon) \geq p_{N_\epsilon} \mathbb{P}(A_\Omega \cap B_\Omega) \geq \left(1 - \frac{1}{2} e^{-\log(1/\epsilon)^d}\right)^2 \geq 1 - e^{-\log(1/\epsilon)^d},$$

where we used $\mathbb{P}(A_\Omega \cap B_\Omega) \geq \mathbb{P}(A_\Omega) + \mathbb{P}(B_\Omega) - 1$. □

We note that the bound given by Eq. (35) on the probability of failure of the algorithm decays super-algebraically with respect to the accuracy $\epsilon$ and is illustrated in Fig. 2 of the main text.

**E. Global error in the spectral norm.** Let $\mathscr{F} : L^2(\Omega) \to L^2(\Omega)$ be the Hilbert–Schmidt integral operator associated with the Green's function of the elliptic operator $\mathcal{L}$ in three dimensions such that

$$\mathscr{F}(f)[x] = \int_\Omega G(x,y) f(y) \, \mathrm{d}y, \quad f \in L^2(\Omega),\, x \in \Omega.$$

Previous works (13, 16) have expressed the error between the learned and exact Green's functions either in the $L^2$-norm or $L^1$-norm by exploiting the connection between the Hilbert–Schmidt norm of the integral operator $\mathscr{F}$ and the $L^2$-norm of the Green's function, since $\|\mathscr{F}\|_{\mathrm{HS}} = \|G\|_{L^2(\Omega \times \Omega)}$. As observed in Section 3C, the peeling algorithm employed in this work introduces a perturbation error on the sketch of the Green's function that depends on the previous hierarchical levels. In particular, relative errors on admissible domains from the past hierarchical levels add up during the procedure. Hence, the algorithm introduced in Section 2, which is based on peeling and the randomized SVD, only guarantees a relative error of $\log(1/\epsilon)^{-1}\epsilon$ on each admissible domain $X \times Y \subset \Omega \times \Omega$, i.e.,

$$\|G - \tilde{G}\|_{L^2(X \times Y)} \leq \log(1/\epsilon)^{-1} \epsilon \|G\|_{L^2(\Omega \times \Omega)}. \tag{42}$$

However, since the number of admissible domains increases exponentially with the hierarchical levels $N_\epsilon$, measuring the error in the $L^2$-norm on the entire domain would significantly deteriorate the error bound, and therefore the number of input-output pairs required to approximate $G$ within $\epsilon$. To circumvent this issue, we exploit the properties of the operator norm of Hilbert–Schmidt operators. In this section, we prove the following proposition, which provides an error bound expressed in the operator norm for the Green's function approximated by Algorithm 1. The proof is deferred to the end of the section.

**Proposition 17** (Approximation error in the HS operator norm). *Let $\mathscr{F} : L^2(\Omega) \to L^2(\Omega)$ be the Hilbert–Schmidt operator associated with the Green's function $G$ of the elliptic operator $\mathcal{L}$ in three dimensions. Let $\tilde{G} : \Omega \times \Omega \to \mathbb{R}$ be the kernel constructed by Algorithm 1 and $\tilde{\mathscr{F}}$ its associated Hilbert–Schmidt integral operator, defined as*

$$\tilde{\mathscr{F}}(f)[x] = \int_\Omega \tilde{G}(x,y) f(y) \, \mathrm{d}y, \quad f \in L^2(\Omega),\, x \in \Omega.$$

*Then, $\|\mathscr{F} - \tilde{\mathscr{F}}\|_2 \leq \epsilon \|\mathscr{F}\|_{\mathrm{HS}}$.*

We remark that Proposition 17 implies that the following equation holds for any $f \in L^2(\Omega)$:

$$\left\| \int_\Omega (G - \tilde{G}) f(y) \, \mathrm{d}y \right\|_{L^2(\Omega)} \leq \epsilon \|G\|_{L^2(\Omega \times \Omega)} \|f\|_{L^2(\Omega)}.$$

Therefore, obtaining a relative error in the operator norm between the exact and learned Green's functions guarantees that the range of the Hilbert-Schmidt operator $\mathscr{F}$, i.e., the solution operator associated with the partial differential operator $\mathcal{L}$, is well approximated. Additionally, expressing the error in the operator norm aligns with the methodology employed by current machine learning techniques for learning solution operators associated with PDEs, such as DeepONets (7, 8) and neural operators (5, 10). In particular, they measure the approximation error of the solution operator on a set of testing pairs. This is equivalent to estimating the error in the spectral norm on the finite dimensional subspace spanned by the $f_j$. We emphasize that measuring the approximation error in the spectral norm is relevant to the setup of state-of-the-art deep learning techniques. We begin by recalling a property for the maximum singular value of an HS operator (analogous to block diagonal matrices).

**Lemma 18** (Operator norm of block diagonal HS operators). *Let $G_1$, $G_2 : \Omega \times \Omega \to \mathbb{R}$ be two kernels such that $\mathrm{Supp}(G_1) \subset D_1 \times D_2$ and $\mathrm{Supp}(G_2) \subset D_3 \times D_4$, where $D_1 \cap D_3 = \emptyset$ and $D_2 \cap D_4 = \emptyset$. Let $\mathscr{F}$, $\mathscr{F}_1$, and $\mathscr{F}_2$ be the Hilbert–Schmidt integral operators associated with the kernels $G_1 + G_2$, $G_1$, and $G_2$, respectively. Then, $\|\mathscr{F}\|_2 = \max\{\|\mathscr{F}_1\|_2, \|\mathscr{F}_2\|_2\}$.*

*Proof.* We start the proof by showing that the first inequality $\|\mathscr{F}\|_2 \geq \max\{\|\mathscr{F}_1\|_2, \|\mathscr{F}_2\|_2\}$ holds:

$$\|\mathscr{F}\|_2 = \sup_{\|f\|_{L^2(\Omega)}=1} \|\mathscr{F}(f)\|_{L^2(\Omega)} \geq \sup_{\substack{\|f\|_{L^2(\Omega)}=1, \\ \operatorname{Supp}(f)\subset D_2}} \|\mathscr{F}(f)\|_{L^2(\Omega)} = \sup_{\|f\|_{L^2(D_2)}=1} \|\mathscr{F}_1(f)\|_{L^2(D_1)} = \|\mathscr{F}_1\|_2,$$

where we identified the operator $\mathscr{F}_1$ with the operator defined on $D_1 \times D_2$, since $\operatorname{Supp}(G_1) \subset D_1 \times D_2$. The same argument applies to $\mathscr{F}_2$ so that $\|\mathscr{F}\|_2 \geq \max\{\|\mathscr{F}_1\|_2, \|\mathscr{F}_2\|_2\}$. For the other inequality, let $f \in L^2(\Omega)$ such that $\|f\|_{L^2(\Omega)} = 1$, and define $f_1 = f|_{D_2}$ and $f_2 = f|_{D_4}$ as the restrictions of $f$ to the domains $D_2$ and $D_4$, respectively. Then,

$$\|f_1\|_{L^2(D_2)}^2 + \|f_2\|_{L^2(D_4)}^2 = \int_{D_2} |f_1(x)|^2\,dx + \int_{D_4} |f_2(x)|^2\,dx \leq \int_\Omega |f(x)|^2\,dx = \|f\|_{L^2(\Omega)}^2 = 1.$$

Therefore, we have

$$\|\mathscr{F}(f)\|_{L^2(\Omega)}^2 = \left\|\int_\Omega (G_1 + G_2)(\cdot, y) f(y)\,dy\right\|_{L^2(\Omega)}^2 = \left\|\int_{D_2} G_1(\cdot, y) f_1(y)\,dy + \int_{D_4} G_2(\cdot, y) f_2(y)\,dy\right\|_{L^2(\Omega)}^2$$

$$= \left\|\int_{D_2} G_1(\cdot, y) f_1(y)\,dy\right\|_{L^2(D_1)}^2 + \left\|\int_{D_4} G_2(\cdot, y) f_2(y)\,dy\right\|_{L^2(D_3)}^2 = \|\mathscr{F}_1(f_1)\|_{L^2(D_1)}^2 + \|\mathscr{F}_2(f_2)\|_{L^2(D_3)}^2$$

$$\leq \|\mathscr{F}_1\|_2^2 \|f_1\|_{L^2(D_2)}^2 + \|\mathscr{F}_2\|_2^2 \|f_2\|_{L^2(D_4)}^2.$$

Taking the supremum of both sides over all possible $f \in L^2(\Omega)$ yields the second inequality as

$$\|\mathscr{F}\|_2^2 \leq \max\{\|\mathscr{F}_1\|_2^2, \|\mathscr{F}_2\|_2^2\}(\|f_1\|_{L^2(D_2)}^2 + \|f_2\|_{L^2(D_4)}^2) \leq \max\{\|\mathscr{F}_1\|_2^2, \|\mathscr{F}_2\|_2^2\}.$$

$\square$

Lemma 18 immediately generalizes to an arbitrary sum of Hilbert–Schmidt integral operators with disjoint kernel supports.

**Corollary 19** (Spectral norm of sums of HS operators). *Let $n \geq 1$ and $G_1, \ldots, G_n : \Omega \times \Omega \to \mathbb{R}$ be $n$ kernels such that $\operatorname{Supp}(G_1) \subset C_1 \times D_1, \ldots, \operatorname{Supp}(G_1) \subset C_n \times D_n$, where $D_i \cap D_j = \emptyset$ and $C_i \cap C_j = \emptyset$ for $1 \leq i \neq j \leq n$. Let $\mathscr{F} : L^2(\Omega) \to L^2(\Omega)$ be the Hilbert–Schmidt integral operator associated with the kernel $\sum_{i=1}^n G_i$, and $\mathscr{F}_i$ be the Hilbert–Schmidt integral operator associated with the kernel $G_i$ for $1 \leq i \leq n$. Then, $\|\mathscr{F}\|_2 = \max_{1 \leq i \leq n} \|\mathscr{F}_i\|_2$.*

*Proof.* The proof follows directly from Lemma 18 by induction. In particular, we consider the HS integral operators associated with the two kernels $H_1 = \sum_{i=1}^{n-1} G_i$ and $H_2 = G_n$ and apply Lemma 18. $\square$

We are now ready to prove Proposition 17.

*Proof of Proposition 17.* Let $\tilde{G} : \Omega \times \Omega \to \mathbb{R}$ be the kernel constructed by Algorithm 1 such that the error with the Green's function $G$ satisfies

$$\|G - \tilde{G}\|_{L^2(X \times Y)} = \begin{cases} \log(1/\epsilon)^{-1} \epsilon \|G\|_{L^2(\Omega \times \Omega)}, & \text{if } X \times Y \text{ is an admissible domain,} \\ \epsilon \|G\|_{L^2(\Omega \times \Omega)}, & \text{otherwise.} \end{cases}$$

We apply Corollary 19 to compute an upper bound on the operator norm of the HS operator associated with the kernel $G - \tilde{G}$. The procedure consists of decomposing the kernel into a sum of kernels over the number of hierarchical levels and colored admissible domains (see Section 2D), such that each element in the sum is a block diagonal kernel whose norm can be estimated by Corollary 19. As an illustration, we assume for simplicity that the kernel $H = G - \tilde{G}$ is written in the weakly admissible hierarchical form with $L = 2$ levels as

$$H = \begin{pmatrix} \begin{array}{cc|cc} \textcolor{red}{H_{11}} & \textcolor{red}{H_{12}} & & \\ \textcolor{red}{H_{12}^\top} & \textcolor{red}{H_{14}} & & \multicolumn{1}{c}{\textcolor{green}{H_2}} \\ \hline & & \textcolor{red}{H_{41}} & \textcolor{red}{H_{42}} \\ \multicolumn{2}{c|}{\textcolor{green}{H_2^\top}} & \textcolor{red}{H_{42}^\top} & \textcolor{red}{H_{44}} \end{array} \end{pmatrix},$$

where the green blocks have HS-norm bounded by $\log(1/\epsilon)^{-1}\epsilon\|G\|_{\mathrm{HS}}$ while the red blocks have HS-norm bounded by $\epsilon\|G\|_{\mathrm{HS}}$. Then we can estimate the operator norm of the associated Hilbert–Schmidt operator by decomposing $H$ as

$$H = \begin{pmatrix} \begin{array}{cc|cc} 0 & 0 & & \\ 0 & 0 & & \multicolumn{1}{c}{\textcolor{green}{H_2}} \\ \hline & & 0 & 0 \\ \multicolumn{2}{c|}{\textcolor{green}{H_2^\top}} & 0 & 0 \end{array} \end{pmatrix} + \begin{pmatrix} \begin{array}{cc|cc} 0 & \textcolor{red}{H_{12}} & & \\ \textcolor{red}{H_{12}^\top} & 0 & & 0 \\ \hline & & 0 & \textcolor{red}{H_{42}} \\ & 0 & & \textcolor{red}{H_{42}^\top} & 0 \end{array} \end{pmatrix} + \begin{pmatrix} \begin{array}{cc|cc} \textcolor{red}{H_{11}} & 0 & & \\ 0 & \textcolor{red}{H_{14}} & & 0 \\ \hline & & \textcolor{red}{H_{41}} & 0 \\ & 0 & & 0 & \textcolor{red}{H_{44}} \end{array} \end{pmatrix}.$$

Hence, the norm of the individual components in the sum are bounded by the norm of each block following Corollary 19, and the norm of the HS operator associated with $H$ is the maximum of the norms of the components. Therefore, in this simple example, we have

$$\|H\|_2 \leq \|H_2\|_2 + \max\{\|H_{12}\|_2, \|H_{42}\|_2\} + \max\{\|H_{11}\|_2, \|H_{14}\|_2, \|H_{41}\|_2, \|H_{44}\|_2\}.$$

In the strongly admissible case with $N_\epsilon \sim \log(1/\epsilon)/(2\log 2)$ levels, we decompose the operator $\mathscr{H} := \mathscr{F} - \tilde{\mathscr{F}}$ as

$$\mathscr{H} = \mathscr{H}_{\text{diag}} + \sum_{L=1}^{N_\epsilon} \sum_{j=1}^{\chi(G_L)} \sum_{i=1}^{n_j} \mathscr{H}|_{X_i^{j,(L)} \times Y_i^{j,(L)}},$$

where $\mathscr{H}_{\text{diag}}$ denote the non-admissible part of the operator $\mathscr{H}$ and $X_i^{j,(L)} \times Y_i^{j,(L)}$ is an admissible domain at level $L$ colored by $1 \leq j \leq \chi(G_L)$. By construction, for a fixed $1 \leq L \leq N_\epsilon$ and $1 \leq j \leq \chi(G_L)$, the restrictions of the kernel of $\mathscr{H}$ to the domains $\{X_i^{j,(L)} \times Y_i^{j,(L)}\}_{1 \leq i \leq n_j}$ have disjoint support. Hence, applying Corollary 19 and using the property that the operator norm is bounded by the HS-norm yield,

$$\|\mathscr{H}\|_2 \leq \|\mathscr{H}_{\text{diag}}\|_2 + \sum_{L=1}^{N_\epsilon} \sum_{j=1}^{\chi(G_L)} \max_{1 \leq i \leq n_j} \{\|\mathscr{H}|_{X_i^{j,(L)} \times Y_i^{j,(L)}}\|_2\} \leq 6^d \epsilon \|\mathscr{F}\|_{\text{HS}} + 6^d \epsilon \|\mathscr{F}\|_{\text{HS}} \leq 2 \times 6^d \epsilon \|\mathscr{F}\|_{\text{HS}}.$$

Here, we used the fact that $N_\epsilon \leq \log(1/\epsilon)$ for sufficiently small $\epsilon$, and $\chi(G_L) \leq 6^d$. Finally, without loss of generality, we can rescale the target accuracy $\epsilon$ in Algorithm 1 to $\tilde{\epsilon} := \epsilon/(2 \times 6^d)$ by increasing the number of input-output pairs by a constant factor, which concludes the proof. □

## Methods

In this section, we describe the deep learning experiments used to generate Fig. 1 of the main text. We compare the performance of three neural network architectures, namely DeepONet (DON) (7), Fourier Neural Operator (FNO) (5), and Green's function learning (GreenLearning) (6), at approximating the solution operator associated with the two-dimensional Poisson equation defined on the domain $\Omega = [0,1]^2$ with homogeneous Dirichlet boundary conditions:

$$-\nabla^2 u = f, \qquad u|_{\partial\Omega} = 0. \tag{43}$$

These methods learn an approximant $\tilde{\mathscr{F}}$ to the solution operator $\mathscr{F}$ associated with Eq. (43), represented by a neural network (NN). Our main motivation is to study the behavior of these PDE learning techniques as the number of training data varies. Hence, we want to understand why specific neural network approaches are data-efficient and others are not. We reproduce the setup of (74) that performs a fair comparison between FNO and DON and provide details of the experiments below. The code is publicly available on GitHub at https://github.com/NBoulle/pde-learning.

**Neural networks architecture.** The FNO architecture is a succession of 4 Fourier layers, which perform convolutions in the Fourier space using the fast Fourier transform (FFT), with the ReLU activation functions. The DON employed in this work has a much larger number of trainable parameters and consists of a product of a "branch network", which is a convolution NN with two 2D convolution layers, and a "trunk network", which is a standard fully connected NN with 4 hidden layers and 128 neurons per layer. GreenLearning enforces prior knowledge of the solution operator associated with Eq. (43) by approximating directly the associated Green's function. Solutions to Eq. (43) are predicted by integrating the NN representation of the Green's function against the forcing terms using a trapezoidal rule. We employ a rational neural network (75) with 4 hidden layers and 50 neurons per layer. Rational NNs offer theoretical and practical advantages over ReLU NNs, as they can capture the singularities of Green's function located along the diagonal (6, 75). DON is implemented in TensorFlow (76) using the DeepXDE library (77), while FNO and GreenLearning are implemented in PyTorch (78). The deep learning experiments were performed on a workstation with a GPU (NVIDIA GeForce RTX 3080 Ti).

**Training and testing datasets.** We train the three neural networks on pairs of training data $\{(f_j, u_j)\}_{j=1}^N$, where we vary $N$ between 2 and 1000. In Fig. 1 of the main text, we report the resulting approximation error $\|\mathscr{F} - \tilde{\mathscr{F}}\|_2$ for $N \leq 200$ to visualize the exponential convergence rate before the plateau due to discretization errors. Here, $\|\mathscr{F} - \tilde{\mathscr{F}}\|_2$ is approximated by the mean relative error on the testing dataset, which consists of $N_{\text{test}} = 200$ pairs of input-output data $\{(f_j^{(\text{test})}, u_j^{(\text{test})})\}_{j=1}^{N_{\text{test}}}$, as

$$\|\mathscr{F} - \tilde{\mathscr{F}}\|_2 \approx \frac{1}{N_{\text{test}}} \sum_{j=1}^{N_{\text{test}}} \frac{\|u_j^{(\text{test})} - \tilde{F}(f_j^{(\text{test})})\|_{L^2([0,1]^2)}}{\|u_j^{(\text{test})}\|_{L^2([0,1]^2)}}.$$

For consistency, we train the different networks ten times and report the mean and standard deviation of the testing errors in Fig. 1. Following (5, 74), the forcing terms are sampled from the Gaussian process $\mathcal{GP}(0, (-\Delta + 9I)^{-2})$ with zero Neumann boundary conditions on the Laplacian. In particular, we sample 1200 forcing terms from the Gaussian process and obtain the associated solutions by solving numerically the Poisson equation. Then, we perform a training/testing split of 1000/200. When we vary the size of the training dataset, we randomly select a portion of the 1000 training pairs to train the models and evaluate them on the 200 testing pairs to avoid train/test contamination. Eq. (43) is solved on a $421 \times 421$ uniform grid, which is then downsampled to a $29 \times 29$ grid when training and testing the NNs.

## References

1. W Hackbusch, *Elliptic differential equations: theory and numerical treatment.* (Springer), 2nd edition, (2017).
2. M Grüter, KO Widman, The Green function for uniformly elliptic equations. *Manuscripta Math.* **37**, 303–342 (1982).
3. T Hsing, R Eubank, *Theoretical foundations of functional data analysis, with an introduction to linear operators.* (John Wiley & Sons), (2015).
4. GE Karniadakis, et al., Physics-informed machine learning. *Nat. Rev. Phys.* **3**, 422–440 (2021).
5. Z Li, et al., Fourier Neural Operator for Parametric Partial Differential Equations in *ICLR.* (2021).
6. N Boullé, CJ Earls, A Townsend, Data-driven discovery of Green's functions with human-understandable deep learning. *Sci. Rep.* **12**, 1–9 (2022).
7. L Lu, P Jin, G Pang, Z Zhang, GE Karniadakis, Learning nonlinear operators via DeepONet based on the universal approximation theorem of operators. *Nat. Mach. Intell.* **3**, 218–229 (2021).
8. S Wang, H Wang, P Perdikaris, Learning the solution operator of parametric partial differential equations with physics-informed DeepONets. *Sci. Adv.* **7**, eabi8605 (2021).
9. CR Gin, DE Shea, SL Brunton, JN Kutz, DeepGreen: Deep learning of Green's functions for nonlinear boundary value problems. *Sci. Rep.* **11**, 1–14 (2021).
10. N Kovachki, et al., Neural operator: Learning maps between function spaces with applications to PDEs. *J. Mach. Learn. Res.* **24**, 1–97 (2023).
11. Z Li, et al., Neural operator: Graph kernel network for partial differential equations. *arXiv preprint arXiv:2003.03485* (2020).
12. S Lanthaler, S Mishra, GE Karniadakis, Error estimates for deeponets: A deep learning framework in infinite dimensions. *Trans. Math. Appl.* **6**, tnac001 (2022).
13. N Boullé, A Townsend, Learning elliptic partial differential equations with randomized linear algebra. *Found. Comput. Math.* pp. 1–31 (2022).
14. MV de Hoop, NB Kovachki, NH Nelsen, AM Stuart, Convergence rates for learning linear operators from noisy data. *SIAM-ASA J. Uncertain. Quantif.* **11**, 480–513 (2023).
15. K Chen, C Wang, H Yang, Deep Operator Learning Lessens the Curse of Dimensionality for PDEs. *arXiv preprint arXiv:2301.12227* (2023).
16. N Boullé, S Kim, T Shi, A Townsend, Learning Green's functions associated with time-dependent partial differential equations. *J. Mach. Learn. Res.* **23**, 1–34 (2022).
17. M Bebendorf, W Hackbusch, Existence of $\mathcal{H}$-matrix approximants to the inverse FE-matrix of elliptic operators with $L^\infty$-coefficients. *Numer. Math.* **95**, 1–28 (2003).
18. F Schäfer, H Owhadi, Sparse recovery of elliptic solvers from matrix-vector products. *arXiv preprint arXiv:2110.05351* (2021).
19. F Schäfer, TJ Sullivan, H Owhadi, Compression, inversion, and approximate PCA of dense kernel matrices at near-linear computational complexity. *Multiscale Model. Simul.* **19**, 688–730 (2021).
20. L Lin, J Lu, L Ying, Fast construction of hierarchical matrix representation from matrix-vector multiplication. *J. Comput. Phys.* **230**, 4071–4087 (2011).
21. PG Martinsson, A fast randomized algorithm for computing a hierarchically semiseparable representation of a matrix. *SIAM J. Mat. Anal. Appl.* **32**, 1251–1274 (2011).
22. J Levitt, PG Martinsson, Linear-complexity black-box randomized compression of hierarchically block separable matrices. *arXiv preprint arXiv:2205.02990* (2022).
23. J Levitt, PG Martinsson, Randomized compression of rank-structured matrices accelerated with graph coloring. *arXiv preprint arXiv:2205.03406* (2022).
24. L Grasedyck, W Hackbusch, Construction and arithmetics of H-matrices. *Computing* **70**, 295–334 (2003).
25. M Bebendorf, *Hierarchical matrices.* (Springer), (2008).
26. W Hackbusch, *Hierarchical matrices: algorithms and analysis.* (Springer), (2015).
27. C Eckart, G Young, The approximation of one matrix by another of lower rank. *Psychometrika* **1**, 211–218 (1936).
28. L Mirsky, Symmetric gauge functions and unitarily invariant norms. *Q. J. Math.* **11**, 50–59 (1960).
29. G Dolzmann, S Müller, Estimates for Green's matrices of elliptic systems by $L^p$ theory. *Manuscripta Math.* **88**, 261–273 (1995).
30. M Bebendorf, Efficient inversion of the Galerkin matrix of general second-order elliptic operators with nonsmooth coefficients. *Math. Comput.* **74**, 1179–1199 (2005).
31. S Börm, L Grasedyck, W Hackbusch, Introduction to hierarchical matrices with applications. *Eng. Anal. Bound. Elem.* **27**, 405–422 (2003).
32. W Hackbusch, A sparse matrix arithmetic based on $\mathcal{H}$-matrices. Part I: Introduction to $\mathcal{H}$-matrices. *Computing* **62**, 89–108 (1999).
33. W Hackbusch, B Khoromskij, SA Sauter, On $\mathcal{H}^2$-Matrices in *Lectures on Applied Mathematics.* (Springer Berlin Heidelberg), pp. 9–29 (2000).
34. PG Martinsson, JA Tropp, Randomized numerical linear algebra: Foundations and algorithms. *Acta Numer.* **29**, 403–572 (2020).
35. N Ailon, B Chazelle, The fast Johnson–Lindenstrauss transform and approximate nearest neighbors. *SIAM J. Comput.*

**39**, 302–322 (2009).
36. KL Clarkson, DP Woodruff, Low-rank approximation and regression in input sparsity time. *J. ACM* **63**, 1–45 (2017).
37. X Meng, MW Mahoney, Low-distortion subspace embeddings in input-sparsity time and applications to robust linear regression in *Proc. STOC*. (ACM), pp. 91–100 (2013).
38. J Nelson, HL Nguyên, OSNAP: Faster numerical linear algebra algorithms via sparser subspace embeddings in *Proc. FOCS*. (IEEE), pp. 117–126 (2013).
39. Y Urano, Master's thesis (New York University) (2013).
40. N Ailon, B Chazelle, Approximate nearest neighbors and the fast Johnson-Lindenstrauss transform in *Proc. STOC*. (ACM), pp. 557–563 (2006).
41. DS Parker, Random Butterfly Transformations with Applications in Computational Linear Algebra, (UCLA), Technical Report CSD-950023 (1995).
42. F Woolfe, E Liberty, V Rokhlin, M Tygert, A fast randomized algorithm for the approximation of matrices. *Appl. Comput. Harmon. Anal.* **25**, 335–366 (2008).
43. N Halko, PG Martinsson, JA Tropp, Finding structure with randomness: Probabilistic algorithms for constructing approximate matrix decompositions. *SIAM Rev.* **53**, 217–288 (2011).
44. N Boullé, A Townsend, A generalization of the randomized singular value decomposition in *ICLR*. (2022).
45. M Lopes, NB Erichson, M Mahoney, Error estimation for sketched SVD via the bootstrap in *ICML*. pp. 6382–6392 (2020).
46. V Rokhlin, A Szlam, M Tygert, A randomized algorithm for principal component analysis. *SIAM J. Mat. Anal. Appl.* **31**, 1100–1124 (2010).
47. AK Saibaba, Randomized subspace iteration: Analysis of canonical angles and unitarily invariant norms. *SIAM J. Mat. Anal. Appl.* **40**, 23–48 (2019).
48. VB Lidskii, On the proper values of a sum and product of symmetric matrices. *Akad. Nauk SSSR* **75**, 769–772 (1950).
49. H Weyl, Das asymptotische verteilungsgesetz der eigenwerte linearer partieller differentialgleichungen (mit einer anwendung auf die theorie der hohlraumstrahlung). *Math. Ann.* **71**, 441–479 (1912).
50. C Davis, WM Kahan, The rotation of eigenvectors by a perturbation. III. *SIAM J. Numer. Anal.* **7**, 1–46 (1970).
51. PÅ Wedin, Perturbation bounds in connection with singular value decomposition. *BIT Numer. Math.* **12**, 99–111 (1972).
52. Y Yu, T Wang, RJ Samworth, A useful variant of the Davis–Kahan theorem for statisticians. *Biometrika* **102**, 315–323 (2015).
53. R Ahlswede, A Winter, Strong converse for identification via quantum channels. *IEEE Trans. Inf. Theory* **48**, 569–579 (2002).
54. R Oliveira, Sums of random Hermitian matrices and an inequality by Rudelson. *Electron. Commun. Probab.* **15**, 203–212 (2010).
55. M Rudelson, Random vectors in the isotropic position. *J. Funct. Anal.* **164**, 60–72 (1999).
56. JA Tropp, User-friendly tail bounds for sums of random matrices. *Found. Comput. Math.* **12**, 389–434 (2012).
57. TT Cai, A Zhang, Rate-optimal perturbation bounds for singular subspaces with applications to high-dimensional statistics. *Ann. Stat.* **46**, 60–89 (2018).
58. E Abbe, J Fan, K Wang, Y Zhong, Entrywise eigenvector analysis of random matrices with low expected rank. *Ann. Stat.* **48**, 1452 (2020).
59. J Cape, M Tang, CE Priebe, Signal-plus-noise matrix models: eigenvector deviations and fluctuations. *Biometrika* **106**, 243–250 (2019).
60. J Cape, M Tang, CE Priebe, The two-to-infinity norm and singular subspace geometry with applications to high-dimensional statistics. *Ann. Stat.* **47**, 2405–2439 (2019).
61. J Eldridge, M Belkin, Y Wang, Unperturbed: spectral analysis beyond Davis-Kahan in *Algorithmic Learning Theory*. (PMLR), pp. 321–358 (2018).
62. J Fan, W Wang, Y Zhong, An $\ell_\infty$ eigenvector perturbation bound and its application to robust covariance estimation. *J. Mach. Learn. Res.* **18**, 1–42 (2018).
63. L Lei, Unified $\ell_{2\to\infty}$ Eigenspace Perturbation Theory for Symmetric Random Matrices. *arXiv preprint arXiv:1909.04798* (2019).
64. Y Zhang, M Tang, Perturbation Analysis of Randomized SVD and its Applications to High-dimensional Statistics. *arXiv preprint arXiv:2203.10262* (2022).
65. YM Chen, XS Chen, W Li, On perturbation bounds for orthogonal projections. *Numer. Algorithms* **73**, 433–444 (2016).
66. G Stewart, JG Sun, *Matrix Perturbation Theory*. (Elsevier), (1990).
67. JG Sun, The stability of orthogonal projections. *J. Graduate Sch. (in Chinese)* **1**, 123–133 (1984).
68. GW Stewart, On the perturbation of pseudo-inverses, projections and linear least squares problems. *SIAM Rev.* **19**, 634–662 (1977).
69. PG Martinsson, Compressing rank-structured matrices via randomized sampling. *SIAM J. Sci. Comput.* **38**, A1959–A1986 (2016).
70. J Mercer, Functions of positive and negative type, and their connection the theory of integral equations. *Proc. R. Soc. A* **209**, 415–446 (1909).
71. A Townsend, LN Trefethen, Continuous analogues of matrix factorizations. *Proc. R. Soc. A* **471**, 20140585 (2015).
72. H Chernoff, A measure of asymptotic efficiency for tests of a hypothesis based on the sum of observations. *Ann. Math.*

*Stat.* pp. 493–507 (1952).
73. RJ Muirhead, *Aspects of multivariate statistical theory*. (John Wiley & Sons), (2009).
74. L Lu, et al., A comprehensive and fair comparison of two neural operators (with practical extensions) based on fair data. *Comput. Methods Appl. Mech. Eng.* **393**, 114778 (2022).
75. N Boullé, Y Nakatsukasa, A Townsend, Rational neural networks in *NeurIPS*. Vol. 33, pp. 14243–14253 (2020).
76. M Abadi, et al., TensorFlow: Large-scale machine learning on heterogeneous systems (2015) Software available from tensorflow.org.
77. L Lu, X Meng, Z Mao, GE Karniadakis, DeepXDE: A deep learning library for solving differential equations. *SIAM Rev.* **63**, 208–228 (2021).
78. A Paszke, et al., Automatic differentiation in PyTorch in *Proc. NIPS Workshops*. (2017).



\end{document}